\pdfoutput=1
\documentclass{article}

\PassOptionsToPackage{numbers, sort}{natbib}
\usepackage[final,main]{neurips_2026}

\usepackage[utf8]{inputenc}
\usepackage[T1]{fontenc}
\usepackage{hyperref}
\hypersetup{pdftitle={Heads That Write, Not Just Point: Image Retrieval Heads in Vision-Language Models}, pdfauthor={Junsung Park, Sangwon Yu, Donghun Kang, Yeongtak Oh, Jaewon Jeong, Gyeongtae Yoo, Han Cheol Moon, Jungbeom Lee, Sungroh Yoon}}
\usepackage{url}
\usepackage{booktabs}
\usepackage{amsfonts}
\usepackage{amsmath}
\usepackage{nicefrac}
\usepackage{microtype}
\usepackage{xcolor}
\usepackage{graphicx}
\usepackage{subcaption}
\usepackage{multirow}
\usepackage{colortbl}
\usepackage{xspace}
\usepackage{tikz}
\usepackage{caption}
\usepackage{multirow}
\usepackage{amsmath, amssymb}
\usepackage{dsfont}
\usepackage{tabularx}

\usetikzlibrary{positioning, arrows.meta, fit, decorations.pathreplacing, calc}

\definecolor{lightblue}{rgb}{0.85,0.92,1.0}
\definecolor{lightred}{rgb}{1.0,0.90,0.90}
\definecolor{deltared}{rgb}{0.70,0.10,0.10}
\definecolor{deltagreen}{rgb}{0.10,0.55,0.10}

\title{Heads That Write, Not Just Point:\\Image Retrieval Heads in Vision-Language Models
}

\author{%
  Junsung Park$^{1}$ \quad Sangwon Yu$^{2}$ \quad Donghun Kang$^{1}$ \quad Yeongtak Oh$^{1}$ \quad Jaewon Jeong$^{3}$ \\
  \bf Gyeongtae Yoo$^{1}$ \quad Han Cheol Moon$^{4}$ \quad Jungbeom Lee$^{5*}$ \quad Sungroh Yoon$^{1,3,6}$\thanks{Corresponding authors: Jungbeom Lee (\texttt{jbeomlee@korea.ac.kr}), Sungroh Yoon (\texttt{sryoon@snu.ac.kr}).} \\[0.6em]
  $^{1}$Department of Electrical and Computer Engineering, Seoul National University \quad
  $^{2}$Upstage AI \\ $^{3}$Interdisciplinary Program in AI, Seoul National University \quad
  $^{4}$Samsung Electronics \\ $^{5}$Korea University \quad
  $^{6}$AIIS, ASRI, INMC, and ISRC, Seoul National University
}

\begin{document}

\maketitle

\begin{abstract}

Large vision-language models (LVLMs) can process interleaved text and multiple images in a single context, but how they internally identify the image relevant to a question remains poorly understood.
We study this grounding step as \emph{in-context image retrieval} and introduce ICIR-MCQ, a controlled probe for head-level analysis of image retrieval.
Using each head's attention mass over image-token spans, we first identify sparse \emph{pointer heads} that reliably point to the target image.
These heads transfer beyond the probe: a single pointer head is competitive with external vision-language retrievers on a naturalistic multi-image retrieval benchmark, showing that LVLMs contain a strong internal image retrieval signal.
Attention alone, however, shows where a head reads from, not what its output passes on to later layers, so we introduce a complementary output-side analysis.
For each head, we train a lightweight image retriever on its value-weighted output and use it to identify \emph{writer heads}, whose outputs contain enough information to identify the target image.
Surprisingly, the pointer and writer criteria select substantially different head sets, with only 35 to 47 heads overlapping among the top-100 heads under each criterion.
Targeted set-partition ablations show that the heads most important for standard multi-image inference are concentrated in this pointer-writer intersection, while heads selected by only one criterion show less consistent separation from random-head controls.

The most consistent causal effects arise from heads selected by both the attention-based pointer and output-side writer criteria.
Code and data: \mbox{\url{https://github.com/parkquasar/Image-Retrieval-Heads}}.
\end{abstract}

\section{Introduction}
\label{sec:intro}
Modern large vision-language models (LVLMs)~\citep{alayrac2022flamingo, wang2025internvl3, bai2025qwen3, li2024llava} can process interleaved text and multiple images in a single context, enabling applications such as document understanding and multi-image question answering.
In multi-image contexts, answering an image-grounded question requires more than visual recognition: the model must first identify which image the question refers to.
We refer to this image-grounding component as \emph{in-context image retrieval}.
This retrieval step gates downstream computation, since grounding the question in the wrong image conditions the answer on the wrong visual evidence.
Recent multi-image benchmarks show that model accuracy drops as the number of candidate images increases, exposing in-context image retrieval as a central behavioral bottleneck~\citep{wu2025visual, wang2024needle}.
However, they leave open how LVLMs internally perform this image selection, and which components in the model carry the image retrieval signal.

Attention-pattern analysis has been a standard tool for identifying specialized heads in transformers~\citep{voita2019analyzing, clark2019does, olsson2022context}.
Recent retrieval-head analyses and attention-based VLM head analyses likewise identify heads by where they attend in the input context~\citep{wu2025retrieval, bi2025unveiling, baek2025large}.
This approach is intuitive and has been effective in practice: attention patterns expose which tokens or spans a head points to.
However, pointing is only one side of the attention operation.
Attention patterns do not by themselves determine what information the head writes into the residual stream.
A head's output is the value-weighted vector produced by mixing value vectors according to its attention weights, and this output, rather than the attention pattern itself, is what downstream layers consume~\citep{kobayashi2020attention}.

We hypothesize that this gap between pointing and writing is especially important for in-context image retrieval in LVLMs, where retrieval is cross-modal: attention points to visual-token spans, but the value-weighted output must carry target-image information in a hidden representation used by later language-generation layers.
This creates a mismatch between pointing and writing.
A head may point cleanly to the target image without writing target-image information in a form recoverable from its output.
Conversely, output-side recoverability and attention-mass strength need not induce the same ranking.
Attention-only head discovery may fail to distinguish heads that merely point to the target image from heads that also write target-image information into the residual stream.

We test this hypothesis with a two-sided analysis of head-level retrieval in LVLMs.
On the input side, we identify \emph{pointer heads} using attention mass over image-token spans (\S\ref{sec:pointer}).
On the output side, we identify \emph{writer heads} by training a lightweight head-specific image retriever on each head's value-weighted output, asking whether the target image can be identified from the information carried by that output (\S\ref{sec:writer}).
We then use targeted ablations to test which head population the model's answer-generation pathway depends on.
Identifying image retrieval heads in LVLMs therefore requires combining where a head points with what it writes.

To carry out this analysis, we introduce ICIR-MCQ (Figure~\ref{fig:cocomcq_sample}; \S\ref{sec:cocomcq}), a controlled probe that gives each sample a clear target image while keeping the task solvable by standard LVLMs.
ICIR-MCQ is designed for controlled head-level discovery rather than as a standalone downstream benchmark; we evaluate transfer and causal relevance on Visual Haystacks~\citep{wu2025visual}, a naturalistic multi-image retrieval benchmark.
Across four LVLMs spanning two families and scales, only 35--47 of the top-100 heads under the pointer and writer criteria overlap.
Targeted set-partition ablations further show that standard multi-image inference depends primarily on this pointer-writer intersection, while heads selected by only one criterion show less consistent separation from random-head controls (\S\ref{sec:validation}).

Our contributions are as follows.

\begin{enumerate}
    \item \textbf{A controlled probe for head-level image retrieval.}
    We introduce ICIR-MCQ, a controlled probe with clear target-image labels for head-level retrieval analysis.

    \item \textbf{Pointer heads as strong implicit image retrievers.}
    We identify sparse \emph{pointer heads} that reproduce at identical (layer, head) indices within shared LLM backbones.
    A single pointer head is competitive with external vision-language retrievers like CLIP and SigLIP.

    \item \textbf{Writer heads via a complementary output-side criterion.}
    We introduce an output-side head discovery method that trains a lightweight image retriever on each head's output, identifying \emph{writer heads} whose outputs retain recoverable target-image information.
    This retriever-based ranking differs substantially from the attention-based pointer ranking.
    
    \item \textbf{Causal validation: retrieval heads are pointers that write.}
    Set-partition ablations show that standard multi-image inference is most sensitive to heads selected by both criteria.

\end{enumerate}
\section{Related Work}
\label{sec:related}

\paragraph{Specialized attention heads and head-level interventions in transformers.}
Prior work has identified specialized transformer heads for syntax~\citep{voita2019analyzing, clark2019does}, induction~\citep{olsson2022context}, and long-context factual retrieval~\citep{wu2025retrieval}.
Many of these analyses identify heads primarily from attention patterns over textual contexts.
However, attention weights alone do not determine the output of an attention layer~\citep{kobayashi2020attention}.
Our output-side analysis is aligned with broader mechanistic work that studies activations, circuits, and interventions beyond raw attention patterns~\citep{elhage2021mathematical, conmy2023towards, li2023inference, todd2024function, geva2023dissecting, baek2026diagnosing}.
We extend this perspective to cross-modal image retrieval in LVLMs, where information about the target image must be written to the residual stream rather than directly emitted as an output token.

\paragraph{Mechanistic interpretability of vision-language models.}
Prior work studies how VLMs integrate visual and textual information across tokens, layers, and attention patterns~\citep{schwettmann2023multimodal, palit2023towards, neo2025towards, zhang2025cross, kang2025see, darcet2024vision}.
Head-level VLM studies have identified visual, OCR, and broad functional heads in single-image, rendered-text, or multimodal VQA settings~\citep{bi2025unveiling, baek2025large, jiang2025investigating, choi2026truth}.
These works either focus on single-image or rendered-text settings, or assign broad functional labels rather than isolating the image-selection mechanism in natural multi-image contexts.
We study natural multi-image retrieval, and our two-sided analysis exposes a discrepancy that attention-based identification alone cannot capture: heads selected by the pointer criterion are not necessarily selected by the writer criterion.

\paragraph{Multi-image vision-language models and benchmarks.}
Recent LVLMs support interleaved text and multi-image inputs~\citep{alayrac2022flamingo, wang2025internvl3, bai2025qwen3, li2024llava}, enabling evaluation over many images in context.
Multi-image benchmarks such as Visual Haystacks and MM-NIAH show that performance degrades as the number of images grows~\citep{wu2025visual, wang2024needle, sharma2024losing, tian2025identifying}, suggesting that target-image grounding is a central bottleneck.
However, they evaluate output behavior rather than the internal retrieval mechanism.
Our work addresses this gap by providing a head-level mechanistic account of in-context image retrieval, using a controlled probe (\S\ref{sec:cocomcq}) for discovery and Visual Haystacks for transfer evaluation (\S\ref{sec:capability}).

\section{Setup}
\label{sec:setup}

\subsection{Preliminaries}
\label{sec:prelim}

\textbf{In-context image retrieval.}
Consider an LVLM processing an interleaved image-text context
\[
    C = (\tau_0, I_1, \tau_1, I_2, \ldots, I_N, \tau_N, u),
\]
where $I_1,\ldots,I_N$ are candidate images, $\tau_0,\ldots,\tau_N$ are intervening text passages, and $u$ denotes task-specific textual instructions, including the question and answer options when applicable.
Answering an image-grounded question in this setting requires grounding the question in the target image before performing the downstream visual computation.
We refer to this image-grounding component, which selects the relevant image within the context, as \emph{in-context image retrieval}.
This retrieval step gates multi-image inference: grounding the query in the wrong image conditions the answer on the wrong visual evidence.
This paper studies the head-level mechanism underlying this retrieval step.

\textbf{Attention notation.}
After tokenization, the interleaved context $C$ becomes a sequence of length $T$ containing both visual and text tokens.
Each image $I_i$ occupies a contiguous visual-token span $[s_i,e_i]$, while different image spans are separated by text-token spans induced by $\tau_0,\ldots,\tau_N$.
For an LVLM with $L$ transformer blocks and $H$ attention heads per block, we denote the post-softmax attention matrix of head $(l,h)$ by $A^{(l,h)} \in \mathbb{R}^{T \times T}$.
For a token position $p$, the attention mass assigned by head $(l,h)$ to image $i$ is
\begin{equation}
    m^{(l,h)}_i(p)
    \;=\;
    \sum_{j=s_i}^{e_i} A^{(l,h)}_{p,j}.
    \label{eq:image_attention_mass}
\end{equation}

Both the input-side and output-side criteria define a per-head, per-sample score vector $s^{(l,h)} \in \mathbb{R}^{N}$ over candidate images, and differ in how they compute it (\S\ref{sec:pointer_criterion} and \S\ref{sec:writer_criterion}).
The score is computed at a task-specified query position $p$, which we instantiate for ICIR-MCQ in \S\ref{sec:cocomcq}.
For a sample with target image $i^\star$, we define the head-level retrieval correctness as
\begin{equation}
    r^{(l,h)}
    \;=\;
    \mathds{1}\!\left[\underset{i}{\mathrm{argmax}}\; s^{(l,h)}_i = i^\star\right].
    \label{eq:r1}
\end{equation}
We report head-level R@1 by averaging this indicator over samples.

\textbf{Models.}
We study four LVLMs spanning two model families and two parameter scales: InternVL3.5-\{4B, 8B\}~\citep{wang2025internvl3} and Qwen3-VL-\{4B, 8B\}~\citep{bai2025qwen3}.
The two model families differ in vision encoder and training data, but the 4B variants share an LLM backbone, as do the 8B variants.
This design lets us examine head-level stability across LLM backbones and, within shared backbones, head-index reproducibility.
We choose these families because they are recent open-weight LVLMs that support interleaved multi-image inputs and allow head-level attention extraction and intervention.
Exact architectural dimensions and implementation details are provided in Appendix~\ref{app:impl}.

\subsection{ICIR-MCQ: A Controlled Probe for Head-Level Image Retrieval}
\label{sec:cocomcq}

\begin{figure}[t]
\centering
\includegraphics[width=\linewidth]{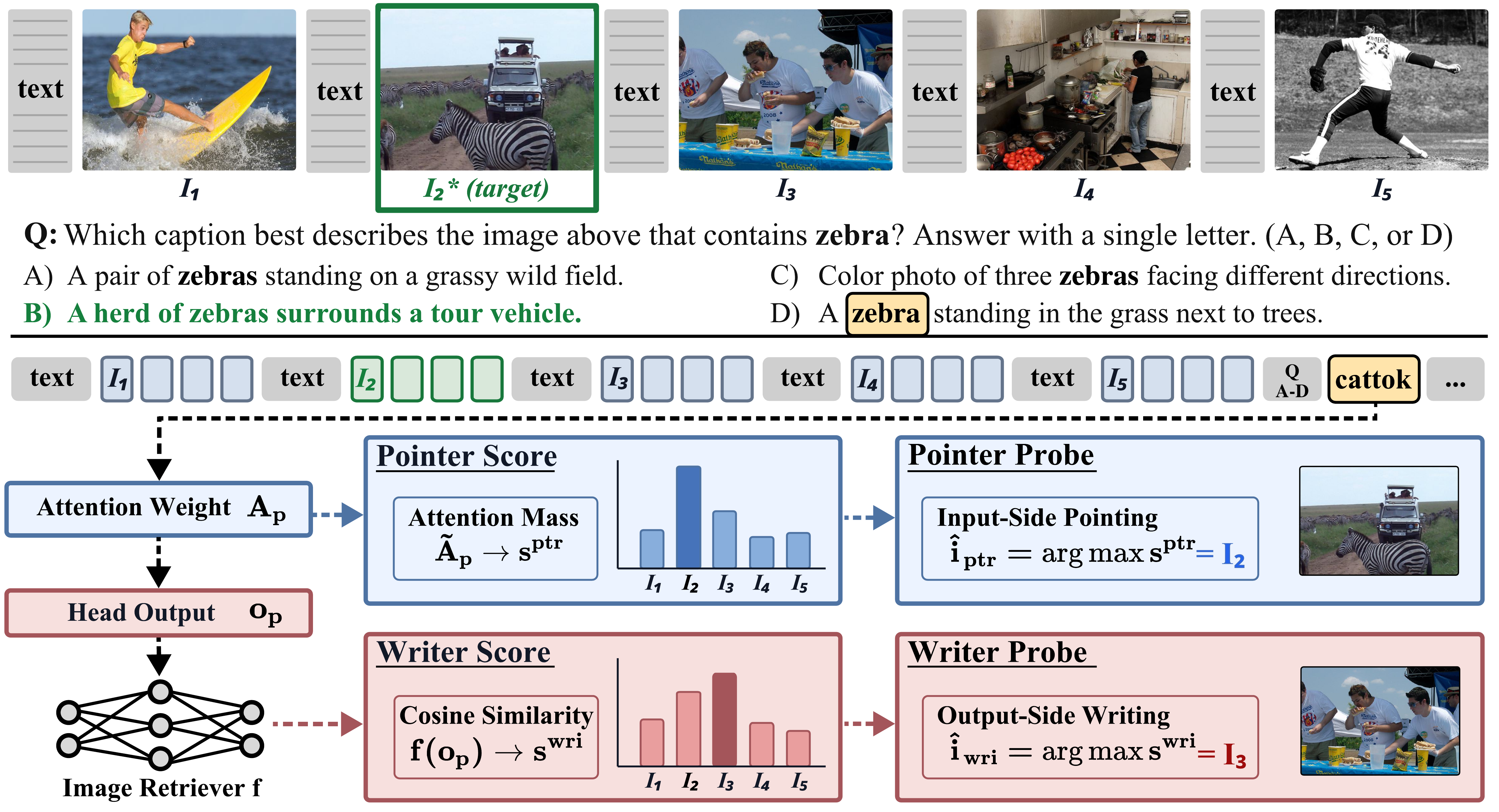}
\caption{An ICIR-MCQ sample. The model selects the caption matching the image containing the queried category (\textit{zebra}). At the \emph{cattok}, the \emph{pointer} criterion ranks images by projection-based attention mass (Eq.~\ref{eq:pointer_score}), while the \emph{writer} criterion ranks images using a retriever trained on the head output. The two criteria can disagree in the same head.}
\label{fig:cocomcq_sample}
\end{figure}

We design ICIR-MCQ as a controlled probe for analyzing image retrieval at the attention head level.
Each sample consists of an interleaved image-text context with $N$ candidate images, a textual query specifying an object category, and four candidate captions.
The model is asked to select the caption that best describes the image containing the queried category (Figure~\ref{fig:cocomcq_sample}, top).
The target image $i^\star$ is the unique image in the context annotated with the queried category; in Figure~\ref{fig:cocomcq_sample}, this is the second image, which contains a \textit{zebra} and matches the caption (B).

ICIR-MCQ is constructed to satisfy three properties.
First, the input is an interleaved image-text context, matching the general form of multi-image inputs to LVLMs.
Second, the intervening text passages do not reveal the target image or the correct caption.
Third, the candidate images are natural images, so head-level findings are not tied to a synthetic visual construction.
To satisfy these requirements, we construct ICIR-MCQ from MS-COCO~\citep{lin2014microsoft}, inheriting the interleaved structure from MM-NIAH's image-retrieval task~\citep{wang2024needle} and substituting natural images for the original images.

More precisely, we keep the inherited text passages and reformulate the task as multiple-choice caption matching.
Because the original MM-NIAH passages are kept fixed while the images and captions are newly substituted from COCO, the passages are independent of the target image and answer options by construction.
We use COCO category annotations to control the per-image attribution target: the queried category is annotated in the target image but not in any distractor image, making the target image unambiguous given the category cue.
Distractor captions are sampled from other COCO images containing the same queried category, so the model cannot solve the task from the category cue alone; it must ground the query in the correct image.

For head-level analysis, we score each head at the last token of the final mention of the queried category, which we call the \emph{cattok}.
This token is a query position at which the category cue has just appeared in text and all candidate image spans are in the causal past.
In Figure~\ref{fig:cocomcq_sample}, the cattok is the last token of \textit{zebra} in option (D).
Scoring at the cattok asks which image each head ranks highest when conditioned on the queried category.

The target image position and correct answer letter are balanced uniformly across options; further sampling and filtering details are in Appendix~\ref{app:cocomcq}.
We compute head-level retrieval R@1 under each criterion's score, defined in \S\ref{sec:pointer_criterion} and \S\ref{sec:writer_criterion}.
ICIR-MCQ serves as our discovery probe and supplies the data for training and evaluating the head-specific image retrievers (\S\ref{sec:writer_criterion}).
\section{Input-Side Retrieval: Pointer Heads}
\label{sec:pointer}

\subsection{Pointer Criterion: Retrieval by Per-Image Attention Mass}
\label{sec:pointer_criterion}
\vspace{2pt}

A natural input-side criterion for image retrieval is whether a head attends to the target image.
We instantiate this as the \emph{pointer criterion}.
For head $(l,h)$ and the cattok position $p$, the score assigned to image $i$ is the attention mass over its visual-token span:
\begin{equation}
    s^{(l,h),\mathrm{ptr}}_i
    \;=\;
    \sum_{j=s_i}^{e_i} \tilde{A}^{(l,h)}_{p,j},
    \qquad
    \tilde{A}^{(l,h)}_{p,j}
    \;=\;
    \mathrm{softmax}\!\left(
    \frac{\mathbf{q}^{(l,h)}_{p} \cdot \mathbf{k}^{(l,h)\top}_{j}}{\sqrt{d_h}}
    \right)_j .
    \label{eq:pointer_score}
\end{equation}
Here $\mathbf{q}^{(l,h)}_p$ and $\mathbf{k}^{(l,h)}_j$ are the direct outputs of the query and key projection layers and the softmax runs over all prefill positions $j$; $\tilde{A}$ is a proxy for the forward-pass attention $A$.
Plugging $s^{(l,h),\mathrm{ptr}}$ into Eq.~\ref{eq:r1} yields the head's R@1 under the pointer criterion.
We call high-R@1 heads \emph{pointer heads}: they rank the target image first by concentrating attention mass on its visual-token span.
The criterion follows the attention-pattern convention used to identify retrieval heads in language-only models~\citep{wu2025retrieval}, adapted here from token-level lookup to span-level image retrieval.

\subsection{Pointer Heads Are Sparse and Localized}
\label{sec:where}
\vspace{1pt}

\begin{table}[!t]
\centering
\small
\setlength{\tabcolsep}{4pt}
\caption{Top-5 pointer heads by head-level attention-mass R@1 on ICIR-MCQ.
The strongest heads cluster in the same mid-block layer band, and within each LLM-backbone group (8B above; 4B below), top heads reproduce at identical (layer, head) indices.
Cells show \texttt{LxxHyy}\,(R@1).}
\label{tab:top_heads_cocomcq}
\resizebox{\linewidth}{!}{%
\begin{tabular}{l ccccc c}
\toprule
Model & Top-1 & Top-2 & Top-3 & Top-4 & Top-5 & Median \\
\midrule
InternVL3.5-8B & \texttt{L21H10}\,(.890) & \texttt{L21H11}\,(.850) & \texttt{L21H08}\,(.842) & \texttt{L20H17}\,(.825) & \texttt{L21H09}\,(.823) & .206 \\
Qwen3-VL-8B   & \texttt{L21H10}\,(.922) & \texttt{L21H08}\,(.883) & \texttt{L21H11}\,(.878) & \texttt{L23H10}\,(.873) & \texttt{L21H09}\,(.869) & .198 \\
\midrule
InternVL3.5-4B & \texttt{L21H10}\,(.866) & \texttt{L21H11}\,(.847) & \texttt{L23H10}\,(.835) & \texttt{L21H08}\,(.826) & \texttt{L21H09}\,(.824) & .217 \\
Qwen3-VL-4B   & \texttt{L21H10}\,(.905) & \texttt{L23H10}\,(.878) & \texttt{L21H11}\,(.870) & \texttt{L21H08}\,(.859) & \texttt{L21H09}\,(.856) & .206 \\
\bottomrule
\end{tabular}}
\vspace{4pt}
\end{table}

Pointer ability is sharply non-uniform: the best pointer head reaches R@1 between .866 and .922 across the four LVLMs (Table~\ref{tab:top_heads_cocomcq}; chance R@1 is .183), while the median head remains near chance.
High-R@1 heads concentrate in a narrow mid-block layer band, with the same band visible on every LVLM; the per-model layer$\times$head R@1 heatmap is reported in Appendix~\ref{app:heads} (Figure~\ref{fig:heatmap}).
We denote by $\mathcal{H}^{\mathrm{ptr}}_K$ the top-$K$ heads under this pointer ranking. Despite differing in vision encoder and training data, LVLMs sharing an LLM backbone show sharp reproducibility at specific (layer, head) indices.
The top-3 pointer heads coincide as a set within each backbone group, and 86 of the top-100 are shared between the 8B models, 80 between the 4B models.
Per-head pointer R@1 grids and within-backbone consistency statistics are reported in Appendix~\ref{app:heads}.

\subsection{Pointer Heads Transfer Beyond the Controlled Probe}
\label{sec:capability}
\vspace{1pt}

We next ask whether pointer heads discovered on ICIR-MCQ transfer to a naturalistic multi-image benchmark.
We use Visual Haystacks~\citep{wu2025visual}: each example contains multiple images and a Yes/No question with an \emph{anchor object} that identifies the target image and a \emph{target object} whose presence in it is queried.
For example, a question may ask, ``In the image that contains a \emph{sofa}, is there a \emph{bottle}?'', where \emph{sofa} is the anchor object and \emph{bottle} is the target object.
For pointer scoring, we use the last token of the anchor-object phrase as the cattok, because this is where the model has enough information to determine which image should be retrieved.
We evaluate at haystack size $N{=}10$, where $N$ denotes the number of candidate images, using each head's per-image attention mass to rank candidate images and measuring whether the top-ranked image is the target image.

\begin{table}[t]
\centering
\caption{Image retrieval R@1 on Visual Haystacks (haystack size $N{=}10$, chance R@1 is $\nicefrac{1}{10}$).
Internal results use the single pointer head \texttt{L21H10} in each LVLM; external retrievers are evaluated on the same samples.
\textbf{IV} = InternVL3.5, \textbf{QV} = Qwen3-VL, and \textbf{QV-Emb} = Qwen3-VL-Embedding-8B.}
\label{tab:vhs_retrieval}
\small
\setlength{\tabcolsep}{4pt}
\begin{tabular}{l@{\hspace{0.6em}}cccc@{\hspace{1.6em}}ccc}
\toprule
\multirow{2}{*}{Method}
& \multicolumn{4}{c}{Single Pointer Head}
& \multicolumn{3}{c}{Retriever Model} \\
\cmidrule(lr){2-5}\cmidrule(lr){6-8}
& IV-8B & QV-8B & IV-4B & QV-4B
& CLIP & SigLIP & QV-Emb \\
\midrule
R@1 & .787 &  .832 & .794 & .836
& .573 & .649 & .751 \\
\bottomrule
\end{tabular}
\vspace{2pt}
\end{table}

\paragraph{Transfer from ICIR-MCQ to a naturalistic benchmark.}
For each LVLM, we use its top pointer head identified on ICIR-MCQ.
Per-image retrieval R@1 ranges from .787 to .836 at haystack size $N{=}10$, far above the chance level of .1 for selecting among ten images.
At this haystack size the 4B variants match or slightly exceed their 8B counterparts, suggesting that strong pointer signals are not confined to the larger models.

\paragraph{A single pointer head is competitive with external retrievers.}
A single attention head from an LVLM not trained with an explicit retrieval objective achieves higher per-image retrieval R@1 (Table~\ref{tab:vhs_retrieval}) than standard external vision-language retrievers---CLIP~\citep{radford2021learning}, SigLIP~\citep{zhai2023sigmoid}, and Qwen3-VL-Embedding-8B~\citep{li2026qwen3}.
This comparison should not be read as a controlled training-objective comparison: the internal pointer head and the external retrievers operate on different representations and conditioning signals.
Rather, it shows that the LVLM contains a highly effective context-conditioned image-selection signal in a single head.
This advantage is consistent with the internal head's access to richer signal: external retrievers compare global image embeddings with context-free text embeddings, whereas the pointer head attends over patch-level keys in the LVLM's own representational space using a query conditioned on the context through the cattok.
Retriever-side work continues to improve both components~\citep{park2025know, kim2026sage}.
Additional retriever details are provided in Appendix~\ref{app:impl}.
\section{Output-Side Retrieval: Writer Heads}
\label{sec:writer}

The pointer criterion asks where a head points, whereas downstream computation uses what it writes to the residual stream.
At the cattok position $p$, the head output is
\begin{equation}
    \mathbf{o}^{(l,h)}_p
    \;=\;
    \sum_{j=1}^{T} A^{(l,h)}_{p,j} \, \mathbf{v}^{(l,h)}_j
    \;\in\;
    \mathbb{R}^{d_h},
    \qquad
    \mathbf{v}^{(l,h)}_j
    \;=\;
    W^{(l,h)}_v \mathbf{x}_j,
    \label{eq:head_output}
\end{equation}
where $\mathbf{x}_j$ is the input to the value projection at position $j$ and $W^{(l,h)}_v$ is the head's value projection.
A head's value projection determines which information from the attended tokens enters the output: even when attention concentrates on the target image, there is no guarantee that the head output contains information sufficient to identify that image.
We therefore introduce a complementary output-side criterion that asks whether the target image can be identified from the head output itself.
This lets us compare attention-based pointing with output-side writing at the same head granularity.

\subsection{Writer Criterion: Retrieval from Head Outputs}
\label{sec:writer_criterion}

We introduce a retriever-based output-side head discovery method.
For each head $(l,h)$, we train a lightweight head-specific \emph{image retriever} on the head output $\mathbf{o}^{(l,h)}_p$.
Following the standard dual-encoder retrieval formulation~\citep{karpukhin2020dense, radford2021learning}, the retriever maps the head output and each candidate image embedding into a shared space and scores candidates by cosine similarity,
\begin{equation}
    s^{(l,h),\mathrm{wri}}_i
    \;=\;
    \tilde{f}_{\theta_{l,h}}(\mathbf{o}^{(l,h)}_p)
    \cdot
    \tilde{g}_{\phi_{l,h}}(\mathbf{e}_i),
    \label{eq:writer_score}
\end{equation}
where $\mathbf{e}_i$ is obtained by mean-pooling candidate image $i$'s projected visual tokens in the LVLM's language-model embedding space, so the writer probe uses the LVLM's own image representation rather than an external encoder.
We train $f_{\theta_{l,h}}$ and $g_{\phi_{l,h}}$ jointly with InfoNCE~\citep{oord2018representation} for each head.

We split ICIR-MCQ into disjoint train/evaluation subsets at the sample level.
For writer-head discovery, we evaluate Eq.~\ref{eq:writer_score} over a held-out image pool of $1{,}200$ candidates per LVLM and compute R@1 by whether the target image ranks first.
We call this metric \emph{writer R@1}, and call high-R@1 heads under this criterion \emph{writer heads}.
Operationally, a writer head is one whose output lets the trained retriever identify the target image, in the spirit of probing intermediate representations~\citep{belinkov2022probing}.
Whether these outputs are used by the model's answer-generation pathway is tested separately in \S\ref{sec:validation}.
Architecture details and hyperparameters are provided in Appendix~\ref{app:learned_retriever}. 

The writer criterion is parallel to the pointer criterion: both produce per-image scores and are evaluated by top-1 retrieval accuracy.
However, pointer R@1 is computed over the images in each prompt, whereas writer R@1 is computed over a held-out image pool.
We therefore use writer R@1 as an output-side ranking criterion, rather than comparing its absolute value directly against pointer R@1.
We denote the top-$K$ pointer and writer heads by $\mathcal{H}^{\mathrm{ptr}}_K$ and $\mathcal{H}^{\mathrm{wri}}_K$.

\subsection{Writer Heads Encode Recoverable Target-Image Information}
\label{sec:writer_results}

The trained retrievers give us a concrete way to test whether a head output carries recoverable target-image information.
Table~\ref{tab:top_writer_heads} shows that the strongest writer heads reach R@1 of $.472$--$.595$ on the held-out pool of $1{,}200$ candidate images, compared with median values of $.077$--$.101$.

\begin{table}[!t]
\centering
\small
\setlength{\tabcolsep}{4pt}
\caption{Top-5 writer heads by writer R@1 on the ICIR-MCQ held-out image pool.
Within each LLM-backbone group (8B above; 4B below), top heads largely overlap.
Writer R@1 is evaluated on a held-out pool of $1{,}200$ candidate images, whereas pointer R@1 in Table~\ref{tab:top_heads_cocomcq} is evaluated on the images in each context; the two R@1 are not directly comparable in absolute scale.
Cells show \texttt{LxxHyy}\,(R@1).}
\label{tab:top_writer_heads}
\resizebox{\linewidth}{!}{%
\begin{tabular}{l ccccc c}
\toprule
Model & Top-1 & Top-2 & Top-3 & Top-4 & Top-5 & Median \\
\midrule
InternVL3.5-8B & \texttt{L24H23}\,(.489) & \texttt{L20H28}\,(.435) & \texttt{L20H15}\,(.426) & \texttt{L21H31}\,(.421) & \texttt{L18H19}\,(.418) & .077 \\
Qwen3-VL-8B   & \texttt{L24H23}\,(.595) & \texttt{L21H17}\,(.470) & \texttt{L26H26}\,(.460) & \texttt{L21H31}\,(.459) & \texttt{L20H28}\,(.458) & .095 \\
\midrule
InternVL3.5-4B & \texttt{L21H31}\,(.472) & \texttt{L20H28}\,(.439) & \texttt{L24H23}\,(.433) & \texttt{L20H15}\,(.428) & \texttt{L21H17}\,(.428) & .080 \\
Qwen3-VL-4B   & \texttt{L24H23}\,(.509) & \texttt{L20H28}\,(.487) & \texttt{L21H31}\,(.486) & \texttt{L21H17}\,(.483) & \texttt{L18H19}\,(.463) & .101 \\
\bottomrule
\end{tabular}}
\vspace{12pt}
\end{table}

\textbf{Writer ability is sparse and mid-block.}
Writer R@1 is sharply non-uniform across heads.
The strongest writer heads concentrate in mid-block layers $18$--$27$ (Appendix~\ref{app:learned_retriever}).
This suggests from the output side that image-retrieval-relevant signals are localized rather than uniformly distributed across the network.
Per-head writer R@1 grids are provided in Appendix~\ref{app:learned_retriever}.

\textbf{Top writer heads reproduce within shared LLM backbones.}
As with pointer heads, writer heads reproduce within shared LLM backbones despite differences in vision encoder and training data.
The 8B models share three of the top-5 writer heads and the 4B models share four, with three heads common to all four models (Table~\ref{tab:top_writer_heads}).
Although writer heads occupy a similar mid-block region as pointer heads, their top-ranked identities differ, motivating a direct comparison between the two rankings.

\subsection{Pointing Is Not Writing}
\label{sec:two_criteria_overlap}

\begin{figure}[t]
\centering
\includegraphics[width=\linewidth]{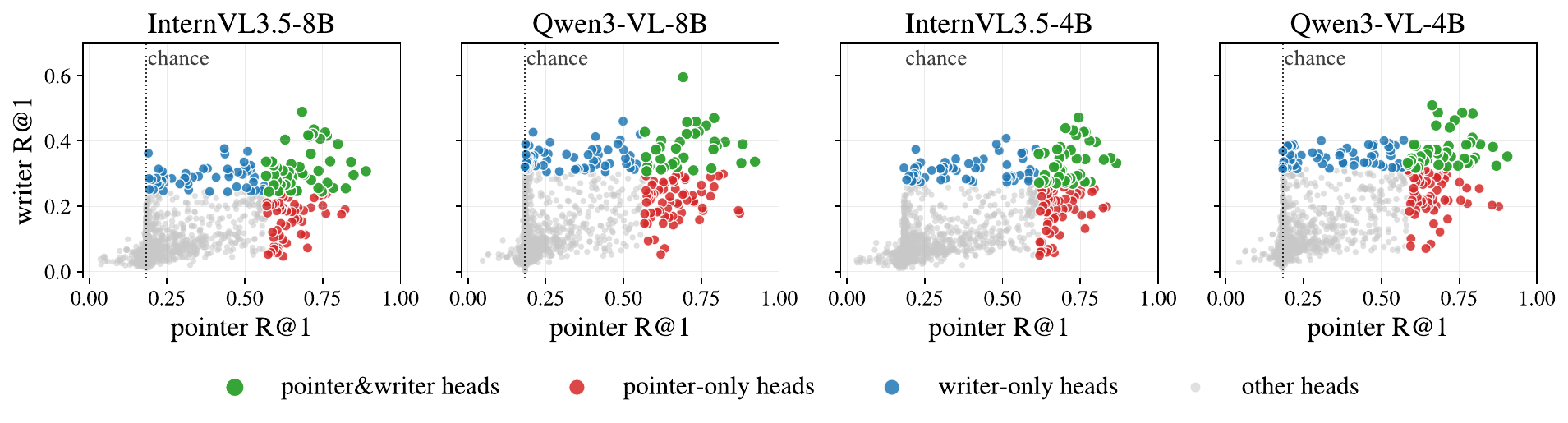}
\caption{Head-level pointer R@1 versus writer R@1 on ICIR-MCQ.
Each point is one attention head: colors denote \emph{pointer\&writer}, \emph{pointer-only}, \emph{writer-only}, and other heads, where membership is defined by the top-100 heads under each ranking.
The vertical dotted line marks pointer-axis chance (R@1 = .183).
The two criteria select substantially different head sets across all four LVLMs.}
\label{fig:pointer_writer_scatter}
\vspace{2pt}
\end{figure}

Pointing and writing identify substantially different heads.
Both rankings concentrate in the same mid-block region, yet at top-100 under each criterion, only 35--47 heads are selected by both.

The off-diagonal heads in Figure~\ref{fig:pointer_writer_scatter} make the dissociation concrete.
Pointer-only heads rank in the pointer top-100 but outside the writer top-100.

Conversely, writer-only heads in $\mathcal{H}^{\mathrm{wri}}_{100} \setminus \mathcal{H}^{\mathrm{ptr}}_{100}$ score highly under the writer criterion despite not being among the strongest pointer heads.
Thus, pointing and writing provide complementary, non-interchangeable rankings.

This dissociation has direct methodological implications: attention-based identification alone cannot distinguish heads that merely point to the target image from heads that also write target-image information into the residual stream.
The remaining question is which population---pointer-only, writer-only, or their intersection---the model's answer-generation pathway actually depends on.
Section~\ref{sec:validation} addresses this with targeted set-partition ablations over these disjoint head sets.
\section{Causal Validation: Retrieval Heads Are Pointers That Write}
\label{sec:validation}

\begin{figure}[t]
\centering
\includegraphics[width=\linewidth]{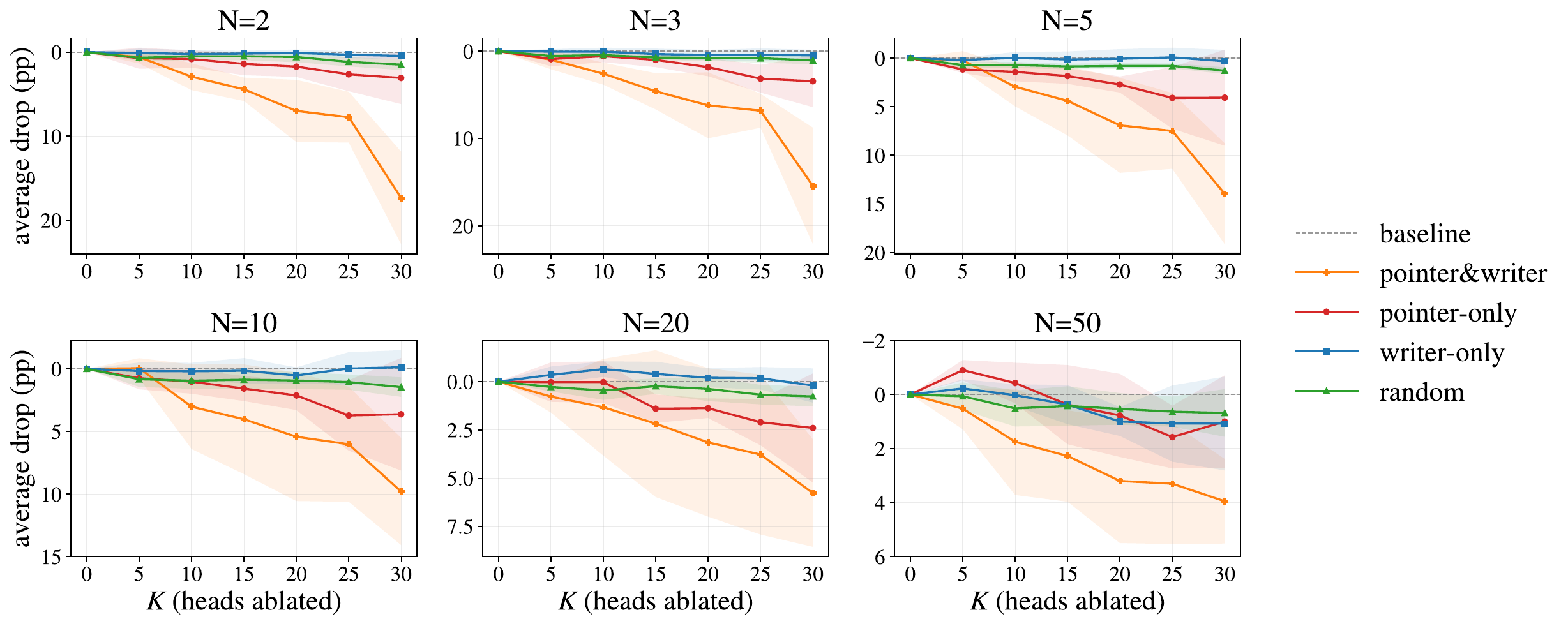}
\caption{Set-partition ablation on Visual Haystacks.
Each panel corresponds to a haystack size $N$, the number of candidate images.
Curves show average accuracy drop relative to the unablated baseline as $K$ heads are ablated.
The pointer-writer overlap shows the most consistent effect. Shaded bands show $\pm 1$ standard deviation across the four LVLMs.}
\label{fig:ablation}
\end{figure}

Sections~\ref{sec:pointer} and~\ref{sec:writer} establish that pointing and writing identify substantially different heads, even though both criteria localize to the same mid-block region.
The key question is which population the standard single-pass forward actually depends on.
If the causally relevant retrieval heads are those that both point to the target image and write recoverable information into the residual stream, the dominant causal contribution should lie in the overlap between the two criteria.
We test this hypothesis with targeted head ablations on Visual Haystacks across multiple haystack sizes, where each example has a ground-truth target image while the model still performs standard multi-image question answering.

\subsection{Ablation Setup}
\label{sec:ablation}

\textbf{Intervention.}
For each selected head $(l,h)$, we zero the head's output contribution before the output projection, thereby removing its additive write to the residual stream.
We use this output-side ablation because it is a stricter test of a head's causal contribution than query-side masking used in prior retrieval-head analysis~\citep{wu2025retrieval}: query masking perturbs where a head attends but can still leave a value-weighted output, whereas our intervention removes the selected head's write itself.
This output-side, head-level zero-ablation is in line with the broader causal-mediation and activation-patching literature for localizing functional roles in transformers~\citep{vig2020investigating, meng2022locating, wang2022interpretability}.

We apply this intervention only over the image-retrieval span of the question.
Following the Visual Haystacks format in Section~\ref{sec:capability}, this span starts at the first token of the anchor object and runs to the last prefill position (the target object ends the question, so only a few format tokens follow it).
The intervention is not applied during the answer generation.
\pagebreak

\textbf{Set partition.}
Directly comparing the pointer top-$K$ and writer top-$K$ sets would confound causal contribution with their partial overlap.
We therefore form the union of the top-100 pointer heads and top-100 writer heads, and partition it into three disjoint subsets.
The \emph{pointer-writer overlap} contains heads selected by both criteria.
The \emph{pointer-only tail} contains heads in the pointer top-100 but not the writer top-100.
The \emph{writer-only tail} contains heads in the writer top-100 but not the pointer top-100.
Across models, the pointer-writer overlap contains 35--47 heads, and each single-criterion tail contains 53 to 65 heads.
We ablate the top-$K$ heads from each subset for $K \in \{5,10,15,20,25,30\}$, with matched-budget random-head controls at the same budgets.
We evaluate at haystack sizes $N \in \{2,3,5,10,20,50\}$.
As $N$ increases, baseline accuracy approaches the Yes/No chance level, reducing the dynamic range available for additional accuracy drops under ablation.

\subsection{The Pointer-Writer Overlap Carries the Causal Effect}
\label{sec:ablation_results}

Ablating the pointer-writer overlap produces the most consistent accuracy drops among the tested head populations (Figure~\ref{fig:ablation}).
The effect generally grows with the ablation budget and is most visible at haystack sizes $2$--$20$, where the pointer-writer curve separates from both single-criterion tails and matched random-head controls.
At larger haystack sizes, the unablated baseline is already closer to the Yes/No chance level, leaving less room for further accuracy drops, so ablation effects are smaller.
Writer-only ablations show little consistent separation from random controls, suggesting that output-side recoverability alone does not identify the heads most relevant to standard inference in this intervention window.
Pointer-only ablations have a modest and less consistent effect, but remain weaker than the pointer-writer overlap in the averaged results for $K \ge 10$.
Overall, the strongest causal evidence points to heads selected by both criteria rather than to heads selected by either criterion alone.
Per-model and per-haystack-size results are provided in Appendix~\ref{app:set_partition}.

\subsection{What the Intervention Isolates}
\label{sec:ablation_interpretation}

The separation from matched random controls, especially at moderate haystack sizes, suggests that the drop is not merely due to removing heads in this sequence region or reducing head capacity in general.
Rather, standard multi-image inference is most sensitive to the particular heads that both point to the target image and write recoverable target-image information.

Together, the pointer criterion reveals strong internal image-selection signals, while the writer criterion shows that attention mass alone does not fully characterize them.
Targeted ablations further show that standard single-pass inference is most sensitive to heads selected by both criteria.
\section{Discussion}
\label{sec:discussion}

\textbf{Why the pointer-writer overlap matters.}
Pointing and writing are controlled by different parts of the attention: Q-K interactions determine where a head points, whereas the value pathway determines what information is written to the residual stream.

Our set-partition ablations show that standard multi-image inference is most sensitive to heads where both properties coincide.
Thus, image retrieval in LVLMs is not fully characterized by where a head attends; the relevant signal must also be written in a form available to downstream computation.

\textbf{From analysis to intervention.}
Our results suggest that interventions based only on amplifying attention to the target image may be insufficient when the selected heads do not write target-image information.
A possible future direction is to modify the value-side pathway of heads that already point well, for example by retraining or steering only a small set of value/output projections while leaving the rest of the model unchanged.
Such targeted interventions could test whether improving the writer side of existing pointer heads can improve multi-image inference without full-model retraining.

\paragraph{Limitations.}
Our analysis focuses on four recent open-weight LVLMs from two model families, and broader model coverage would further clarify how general the pointer-writer structure is.
Our naturalistic evaluation isolates single-target image retrieval in Visual Haystacks; extending the framework to cross-image comparison, multi-target grounding, and compositional reasoning is a natural direction for future work.
Complementary generation-time interventions could further characterize how retrieved visual information is converted into the final textual answer.
\section{Conclusion}
\label{sec:conclusion}
We studied in-context image retrieval in LVLMs at the attention head level.
Using ICIR-MCQ, we identified sparse \emph{pointer heads} whose per-image attention mass ranks the target image first.
These heads localize to a consistent mid-block region and transfer beyond the controlled probe, providing a strong input-side retrieval signal.
However, attention mass alone is incomplete: a head selected by the pointer criterion is not necessarily selected by the writer criterion.
We therefore introduced a retriever-based output-side \emph{writer} criterion and found that pointer and writer rankings identify related but distinct head sets.
Targeted set-partition ablations show that standard single-pass inference is most sensitive to heads selected by both criteria rather than to either single-criterion tail.
Together, these results support a two-sided account of image retrieval in LVLMs: our strongest causal evidence favors heads selected by both the attention-based pointer and output-side writer criteria.

\begin{ack}
This work was supported by the National Research Foundation of Korea (NRF) grants funded by the Korea government (MSIT) (Nos.\ 2022R1A3B1077720, 2022R1A5A7083908, and RS-2026-25488668); by the Institute of Information \& Communications Technology Planning \& Evaluation (IITP) grants funded by the Korea government (MSIT) (No.\ RS-2021-II211343, Artificial Intelligence Graduate School Program, Seoul National University; No.\ RS-2022-II220959; No.\ IITP-2026-RS-2024-00397085, Leading Generative AI Human Resources Development; and No.\ IITP-2026-RS-2020-II201819, IITP-ICT Creative Consilience Program); and by the BK21 FOUR program of the Education and Research Program for Future ICT Pioneers, Seoul National University in 2026.
This research was also supported by the ANCHOR through the Seoul ANCHOR Center, funded by the Ministry of Education (MOE) and the Seoul Metropolitan Government (2026-ANCHOR-01-003-09), and by Samsung Electronics Co., Ltd (IO260506-16457-01).
\end{ack}

\bibliographystyle{plainnat}
\bibliography{refs}

\appendix

\clearpage
\section{Implementation Details}
\label{app:impl}

\paragraph{Model shape.}
The four LVLMs share the same language-model shape: 36 transformer blocks with 32 attention heads each, $L \cdot H = 1{,}152$ heads in total, and head dimension $d_h = 128$. Each model is loaded from its HuggingFace snapshot in bfloat16 and run with FlashAttention-2~\citep{dao2024flashattention}.

\paragraph{Attention score and head-output capture.}
We register forward hooks on \texttt{q\_proj} and \texttt{k\_proj} of each transformer block to capture the query and key projections needed to recompute per-head attention scores during prefill, and a forward-pre-hook on \texttt{o\_proj} to capture each head's output (the value-weighted sum entering the output projection) for the writer probe.
For models with Grouped Query Attention~\citep{ainslie2023gqa}, we expand KV heads to match the query heads before computing per-head attention.
The pointer score (Eq.~\ref{eq:pointer_score}) is computed from these captured projections at the designated readout row only; the forward pass itself is unchanged.

\paragraph{Visual Haystacks setup.}
For each haystack size, a fixed random seed (42) controls filler-image selection and haystack shuffling, so that every model sees the same haystack composition and order.

\paragraph{Image preprocessing and per-image input budget.}
We fix the per-image input budget at $448 \times 448$ pixels for all four LVLMs we study so that cross-model comparisons are not confounded by input size.
\textbf{InternVL3.5} uses a $448 \times 448$ resize (BICUBIC) followed by ImageNet normalization, yielding $256$ visual tokens per image (the architecture's native per-tile budget).
\textbf{Qwen3-VL} uses an aspect-ratio-preserving resize within a configurable pixel budget by default; we override the processor with \texttt{min\_pixels = max\_pixels = }$448 \times 448$, which fixes the pixel budget rather than the resolution: the aspect ratio is preserved and each side is rounded to a multiple of $32$, giving roughly $192$ visual tokens per image, about three quarters of InternVL3.5's $256$.
The same transforms are used in baseline and ablation runs, so within-model comparisons are unaffected by the token-count difference between the two families.

\paragraph{Generation configuration.}
All forward passes run in bfloat16.
Decoding is greedy (\texttt{do\_sample=False}, \texttt{num\_beams=1}) with \texttt{max\_new\_tokens=128}, so any randomness in answers is removed.
For VHs (Yes/No format), generation stops at the first ``yes'' or ``no'' word, with no effect on parsed accuracy.

\paragraph{External retriever evaluation.}
We evaluate CLIP ViT-L/14, SigLIP so400m-patch14-384, and Qwen3-VL-Embedding-8B as external retrievers on Visual Haystacks.
For CLIP and SigLIP, we follow the standard OpenAI 80-template prompt ensemble~\citep{radford2021learning}: the text embedding for the anchor object is the mean of the $\ell_2$-normalized embeddings of the 80 templated prompts, and the per-image score is the cosine similarity between this ensembled text embedding and the image embedding.
The anchor noun is inserted without its article, and SigLIP prompts are padded to the 64-token maximum length used in its pretraining, which its text encoder requires.
For Qwen3-VL-Embedding-8B we use a natural retrieval phrasing query, ``an image with \{anchor\}'', and score images by cosine similarity to this query embedding.
In all three retrievers we select the argmax image as the predicted target.

\paragraph{Ablation procedure.}
For the targeted ablation experiments (\S\ref{sec:validation}), a forward pre-hook on \texttt{o\_proj} zeroes the selected head's slice of the \texttt{o\_proj} input over the image-retrieval span: from the first token of the anchor-object phrase to the last prefill position (the target-object phrase ends the question, so the remaining covered tokens are the fixed answer-format suffix).
 The first answer token is decoded from the last prefill position, where the ablation is already in effect (Appendix~\ref{app:gen_window} checks that continuing the intervention through generation does not change the conclusions).
Random selection draws $K$ heads uniformly from the $1{,}152$ heads at the matched budget (averaged over three seeds in the set-partition ablation of \S\ref{sec:ablation}); the unablated baseline runs the same pipeline with no head zeroed.

\paragraph{Computational resources.}
All main-text experiments run on NVIDIA L40 GPUs (48GB VRAM).
Single-GPU inference suffices for all four LVLMs at the haystack sizes evaluated in this paper ($N \in \{2, 3, 5, 10, 20, 50\}$).
A Visual Haystacks pass at $N{=}10$ takes about 1.5\,s per sample for InternVL3.5-8B and about 4\,s for Qwen3-VL-8B, whose answers are longer, and head-output extraction on ICIR-MCQ about 13\,s per sample; the set-partition grid of \S\ref{sec:validation} comprises about 220 cells of 1,000 samples per LVLM, on the order of a few hundred GPU-hours per LVLM.

\section{ICIR-MCQ: Additional Construction Details}
\label{app:cocomcq}

ICIR-MCQ is described in the main text (\S\ref{sec:cocomcq}) and used as the head-discovery benchmark throughout \S\ref{sec:pointer} and \S\ref{sec:writer}.
This appendix gives the design rationale, the sampling procedure, the split sizes, and one abridged prompt.

\paragraph{Why this design (extended).}
Three properties matter for head-level attribution.
First, the answer is a single discrete token whose correctness is unambiguous.
Second, all four caption options mention the queried category (the three distractor options are captions of other images of that category), so the answer cannot be identified from the category cue alone; the distractor images are drawn from other categories, so the target image is unambiguous given the cue.
Third, the small candidate set ($N \leq 6$) keeps the head-level R@1 estimator stable at the 6{,}000-sample scale, allowing a full sweep of all $L \cdot H{=}1{,}152$ heads without budgeted subsetting.

\paragraph{Sampling procedure.}
We draw each sample by (i) selecting a target image and one of its own COCO captions that mentions the queried category (sampling from the union of the MS-COCO 2017 train and validation splits), (ii) selecting $N_{\text{choices}}-1$ distractor images from other COCO categories, excluding any image annotated with the target's category, so that the queried category is annotated in the target image alone, and (iii) constructing a four-way multiple choice over caption options A--D, in which one option is the target's own caption and the other three are captions of other images of the queried category, so that all four options mention the category.
We constrain (a) the target image's position among the $N_{\text{choices}}$ images to be uniform across the $N_{\text{choices}}$ slots, and (b) the answer letter (the slot of the target's caption among the four options A--D) to be uniformly distributed across $\{$A, B, C, D$\}$.
The number of MCQ option letters (always four) is independent of $N_{\text{choices}}$; the candidate images themselves are unlabeled and identified by their position in the prompt rather than by a letter.

\paragraph{Split sizes.}
We release two splits: $N_{\text{choices}}{=}5$ with 3{,}000 samples and $N_{\text{choices}}{=}6$ with 3{,}000 samples, for a total of 6{,}000.

The released file, its data card, and the builder are available at \url{https://huggingface.co/datasets/jerryray/ICIR-MCQ}; the analysis code is at \url{https://github.com/parkquasar/Image-Retrieval-Heads}.

\paragraph{Robustness of head rankings.}
The top head is invariant between the two $N_{\text{choices}}$ splits for both 8B models (L21H10 in both cases), and the top-5 sets coincide across the splits, with only the ordering of ranks 2--5 differing.

\subsection[An abridged ICIR-MCQ prompt]{An abridged ICIR-MCQ prompt}
\label{app:prompt_example}
Figure~\ref{fig:cocomcq_sample} is a shortened schematic of another sample of the same category.
Below is the prompt of one actual sample (sample 4860 of the released file; six candidate images), with each text passage cut to its first sentence or so.
Passages are the interleaved documents inherited from MM-NIAH (\S\ref{sec:cocomcq}) and are unrelated to the images; the images are inserted at the \texttt{<image>} markers in order.
The category is named in the question and again in each of the four options, which describe different images of that category, so the last mention of the category (the cattok) falls inside the options (option D here).

\begin{small}
\begin{verbatim}
What the real geeks soon realize, however, is that it's fun to brew with the
  Aeropress. You can literally try thousands of different recipes and still
  [...]
<image 1>
"You said an interesting word, 'rumor.' I've been in the business [...]
<image 2>
Here is the equivalent circuit diagram:The below code will demonstrate how it
  works by turning the motor on and off. [...]
<image 3>
The below circuit can be used to demonstrate. Note that there are no external
  components at all. [...]
<image 4>
It includes one one-bedroom unit, and 11 two-bedroom units. The property also
  has parking, controlled access entry and a laundry facility. [...]
<image 5>
Weirick will specialize in working with landlords and tenants regarding their
  industrial real estate needs in the San Fernando Valley. [...]
<image 6>
CDC's latest plan, which was accepted, was to remove the toxic soil and pile
  it up on three separate berms on the property. [...]
Which caption best describes the image above that contains zebra? Answer
  with a single letter (A, B, C, or D).
A) A group of zebras and wildebeest grazing in a field.
B) The two zebra stand next to each other in the wilderness.
C) A group of zebras and other animals grazing in a field with a rainbow in
  the background.
D) The zebra eats the grass near a building.
\end{verbatim}
\end{small}
Target image: position 4 of 6. Correct option: A.

\section{Pointer Heads: Extended Results}
\label{app:heads}

This appendix collects pointer-head results that are summarized in the main text: the per-LVLM head R@1 grid, pairwise ranking similarity across the four LVLMs, the per-haystack R@1 of the top pointer head on Visual Haystacks, and that head's R@1 next to the model's own accuracy.

\subsection{Per-LVLM head R@1 grid}
\label{app:heads_heatmap}

\begin{figure}[h]
\centering
\includegraphics[width=\linewidth]{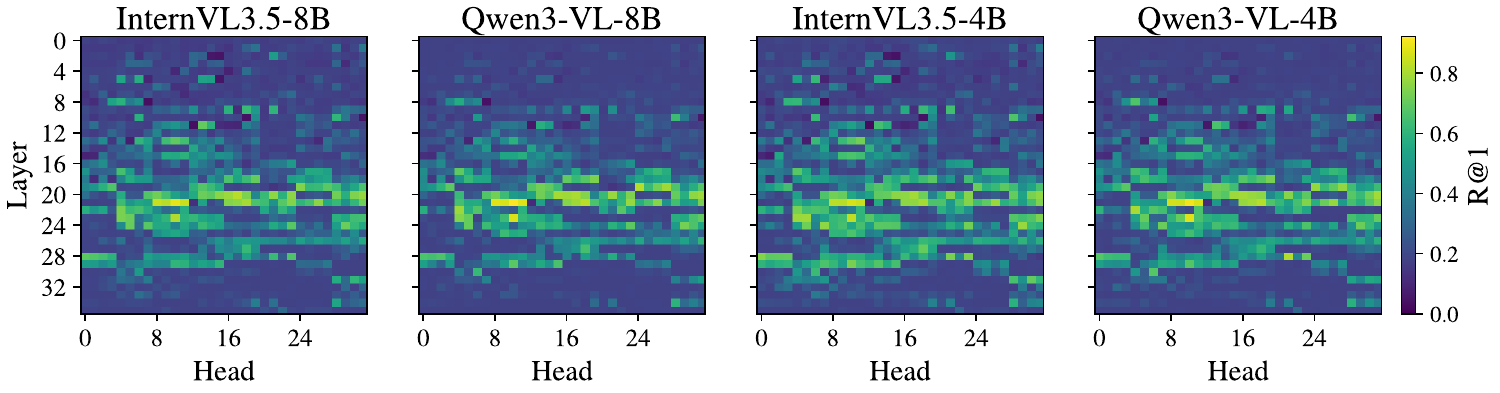}
\caption{Head-level attention-mass R@1 on ICIR-MCQ for all four LVLMs, shown as layers $\times$ heads with a shared color scale.
Pointer ability concentrates in a narrow mid-block layer band, and the same band appears across both model families and both parameter scales.
The strongest pointer head per model is \texttt{L21H10} on every panel (Table~\ref{tab:top_heads_cocomcq}).}
\label{fig:heatmap}
\end{figure}

Figure~\ref{fig:heatmap} shows the pointer R@1 of every head in each LVLM.

\subsection{Pairwise ranking similarity across the four LVLMs}
\label{app:heads_pairwise}

Table~\ref{tab:pairwise_similarity} reports pairwise statistics on the per-head R@1 ranking across the four LVLMs.
For every pair of models it reports four statistics on the per-head R@1 ranking, defined in the caption.
Pairs are grouped into shared-LM-backbone (same LM scale across the two families) and cross-LM-backbone (different LM scale).

\begin{table}[h]
\centering
\small
\setlength{\tabcolsep}{4pt}
\caption{Pairwise per-head R@1 similarity across the four LVLMs.
Pearson and Spearman are computed on the full $1{,}152$-d head ranking; layer-max correlation is on the per-layer maximum R@1 vector (length 36); top-$K$ overlap is $|\mathcal{H}^{\mathrm{ptr}}_K(\text{model 1}) \cap \mathcal{H}^{\mathrm{ptr}}_K(\text{model 2})| / K$.
The top-1 pointer head is \texttt{L21H10} on every model.}
\label{tab:pairwise_similarity}
\begin{tabular}{l cccc cc}
\toprule
& Pearson & Spearman & LayerMax & Top-3 & Top-10 & Top-100 \\
\midrule
\multicolumn{7}{l}{\emph{Shared-LM-backbone pairs}} \\
InternVL3.5-8B $\leftrightarrow$ Qwen3-VL-8B & 0.965 & 0.922 & 0.921 & 1.00 & 0.70 & 0.86 \\
InternVL3.5-4B $\leftrightarrow$ Qwen3-VL-4B & 0.966 & 0.927 & 0.933 & 1.00 & 0.60 & 0.80 \\
\midrule
\multicolumn{7}{l}{\emph{Cross-LM-backbone pairs}} \\
InternVL3.5-8B $\leftrightarrow$ InternVL3.5-4B & 0.977 & 0.946 & 0.962 & 0.67 & 0.70 & 0.85 \\
Qwen3-VL-8B    $\leftrightarrow$ Qwen3-VL-4B    & 0.971 & 0.939 & 0.980 & 0.67 & 0.60 & 0.85 \\
InternVL3.5-8B $\leftrightarrow$ Qwen3-VL-4B    & 0.964 & 0.916 & 0.919 & 0.67 & 0.70 & 0.82 \\
InternVL3.5-4B $\leftrightarrow$ Qwen3-VL-8B    & 0.937 & 0.900 & 0.932 & 0.67 & 0.70 & 0.75 \\
\bottomrule
\end{tabular}
\end{table}

Rankings are highly aligned for every pair (Spearman $\ge 0.90$, top-100 overlap $\ge 0.75$); the top-3 set is identical within each shared-LM-backbone pair and shares 2 of 3 heads across backbones.

\subsection{Per-haystack R@1 decay}

Table~\ref{tab:vhs_r1_by_hs_app} reports the per-haystack R@1 of \texttt{L21H10}, the top ICIR-MCQ pointer head on every model, on Visual Haystacks.
Ranked by VHs R@1 at $N{=}10$, it is also the top head on both 8B models and ranks second on InternVL3.5-4B and third on Qwen3-VL-4B.
R@1 generally falls with haystack size, and the two scales within a family are not consistently ordered: the InternVL3.5 pair stays within $0.8$\,pp at every haystack size, while Qwen3-VL-4B trails Qwen3-VL-8B by up to $6.5$\,pp at $N \in \{2, 3, 20\}$ and matches it at $N \in \{5, 10\}$.
The pointer head is strong at both scales, in line with the cross-scale stability noted in \S\ref{sec:where}, but R@1 itself is not ordered by model scale.

\begin{table}[!htbp]
\centering
\caption{R@1 of the shared top pointer head \texttt{L21H10} on VHs across haystack sizes $\in\{2,3,5,10,20\}$ ($n{=}1{,}000$ per cell). Cattok-position attention argmax accuracy. $N{=}50$ is not included in this readout.}
\label{tab:vhs_r1_by_hs_app}
\begin{tabular}{lccccc}
\toprule
Haystack size & 2 & 3 & 5 & 10 & 20 \\
\midrule
InternVL3.5-8B & .947 & .914 & .860 & .787 & .677 \\
Qwen3-VL-8B   & .959 & .931 & .901 & .832 & .782 \\
InternVL3.5-4B & .939 & .907 & .859 & .794 & .682 \\
Qwen3-VL-4B   & .894 & .901 & .895 & .836 & .745 \\
\bottomrule
\end{tabular}
\end{table}

\subsection{Single-head retrieval versus the model's own accuracy}
\label{app:qa_table}
Table~\ref{tab:qa_table} places the top pointer head's retrieval R@1 next to the accuracy of the unmodified model on the same inputs, for both benchmarks.
On ICIR-MCQ the model answers with a letter and we score the restricted argmax over the four option tokens at the first answer position, on all 6,000 samples.
On Visual Haystacks ($N = 10$) accuracy is the Yes/No accuracy of the unablated baseline used in \S\ref{sec:validation}.
On ICIR-MCQ the head's R@1 and the model's accuracy are within a few points of each other, whereas on Visual Haystacks the single head identifies the target image 11 to 16 points more often than the model answers the question correctly.

\begin{table}[h]
\centering
\small
\caption{Top pointer head (L21H10) retrieval R@1 and base-model accuracy on ICIR-MCQ (6,000 samples) and Visual Haystacks ($N = 10$, 1,000 samples).}
\label{tab:qa_table}
\begin{tabular}{l c c c c}
\toprule
& \multicolumn{2}{c}{ICIR-MCQ} & \multicolumn{2}{c}{Visual Haystacks} \\
\cmidrule(lr){2-3} \cmidrule(lr){4-5}
Model & Head R@1 & Model acc.\ (\%) & Head R@1 & Model acc.\ (\%) \\
\midrule
InternVL3.5-4B & .866 & 85.5 & .794 & 68.1 \\
InternVL3.5-8B & .890 & 88.9 & .787 & 65.9 \\
Qwen3-VL-4B & .905 & 90.1 & .836 & 67.5 \\
Qwen3-VL-8B & .922 & 91.3 & .832 & 67.6 \\
\bottomrule
\end{tabular}
\end{table}

\section{Writer probe: training details and extended agreement analysis}
\label{app:learned_retriever}

This appendix gives the training details of the writer probe (\S\ref{sec:writer}) and compares the writer and pointer rankings; Appendices~\ref{app:writer_controls} and~\ref{app:position} report control experiments on the candidate pool, on object regions, and on the readout position.

\paragraph{Head-level training setup.}
For each attention head, we treat its attention-value-weighted output at the query position as a $d_h$-dimensional image embedding and train a small head-level projection with a contrastive (InfoNCE-style) objective using ICIR-MCQ as positive-negative pairs.
Reading head-level representations through a small trained probe follows the inference-time-intervention~\citep{li2023inference} and function-vector~\citep{todd2024function} setups; we adapt the construction to a cross-modal retrieval objective.
The probe projects the head's output and the candidate image embedding into a shared $256$-dimensional space through one linear layer each, followed by L2 normalization, and scores candidates by dot product.
We selected this configuration (one projector layer, L2 normalization, batch 256, learning rate $1\!\times\!10^{-3}$) from a sweep over architecture, projector depth, batch size, learning rate, and normalization on the 1{,}200-example evaluation split used for checkpoint selection and the reported ranking; the Qwen3-VL-8B probe uses batch 128.
The LVLM weights are not modified; only the head-level probe is trained.

\begin{figure}[h]
\centering
\includegraphics[width=\linewidth]{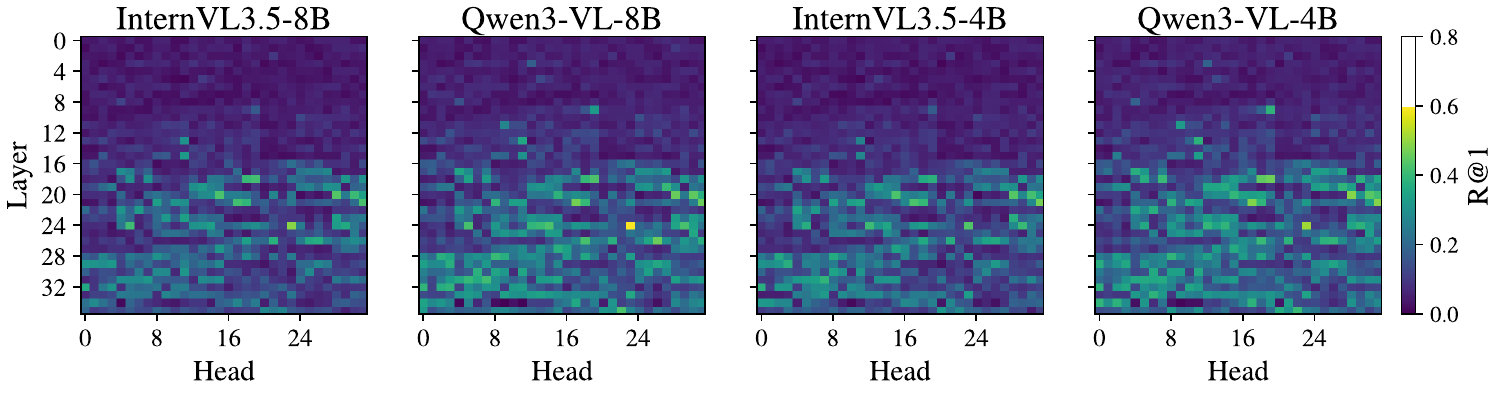}
\caption{Head-level writer R@1 on the ICIR-MCQ held-out evaluation pool (1{,}200 candidate images; chance R@1 $\approx .001$) for all four LVLMs, shown as layers $\times$ heads with a shared color scale.
Writer R@1 concentrates in a narrow mid-block layer band, the same region picked out by the pointer ranking (Figure~\ref{fig:heatmap}).}
\label{fig:writer_heatmap}
\end{figure}

\paragraph{Agreement with the pointer ranking.}
The two criteria both concentrate in mid-block layers, but not on the same layers: the pointer ranking peaks at layer 21 on all four LVLMs, whereas the strongest writer heads sit at layer 24 on three of the four (layer 21 on InternVL3.5-4B). The writer top-100 is also more layer-dispersed than the pointer top-100 (47--60 versus 71--82 of 100 heads inside layers 18--27). The two rankings likewise disagree on which specific heads are top-ranked; the writer top-5 per LVLM are reported in Table~\ref{tab:top_writer_heads} and do not match the pointer top-5 of Table~\ref{tab:top_heads_cocomcq}.
Figure~\ref{fig:writer_heatmap} shows the full per-head writer R@1 grid, mirroring the pointer-side grid in Figure~\ref{fig:heatmap}: the bright band sits in the same mid-block region in both rankings, with the writer band extending slightly further toward later layers.
Ranks diverge more in the mid-tail than at the top.

\paragraph{Writer-ranked vs.\ pointer-ranked ablation.}
Whether the heads that only one criterion selects are causally used is tested by the set-partition ablation (\S\ref{sec:ablation}, Appendix~\ref{app:set_partition}): at $K \ge 10$ the overlap of the two top-100 sets produces the largest drops on average across models, the writer-only tail leaves accuracy near the random control on all four LVLMs, and the pointer-only tail is model-dependent (on InternVL3.5-8B it drops as much as or more than the overlap at $N \ge 5$, $K \ge 25$; near the control on Qwen3-VL-8B).

\section{Writer Criterion: Object-Region and In-Context Controls}
\label{app:writer_controls}

\subsection{Object regions instead of full images}
The writer probe (\S\ref{sec:writer}) retrieves the embedding of the whole target image, so a head could in principle rank well by carrying global image identity rather than information about the queried object.
We therefore retrained the per-head retriever with the target represented by the embedding of the object region, the COCO box of the queried category inside the target image, with the probe configuration of Appendix~\ref{app:learned_retriever}.
As a control for training variance we also retrained the full-image probe once more under the same settings (matched control).
Table~\ref{tab:object_region} compares the resulting head rankings with the full-image writer ranking of the main text, and reports a same-category retrieval test in which the candidate pool contains only object regions of the queried category taken from other images, so that category identity alone cannot solve the task; for a query with $m_i$ same-category distractor regions chance is $1/(m_i+1)$, averaged over queries.

\begin{table}[h]
\centering
\small
\setlength{\tabcolsep}{3.5pt}
\caption{Object-region writer probes. Spearman $\rho$ of the object-region head ranking against the full-image writer ranking (all 1,152 heads); $\rho$ and top-100 overlap of a retrained full-image control against the same ranking; same-category retrieval with the mean number $m$ of same-category distractor regions, the per-query chance $1/(m_i+1)$ averaged over queries, and mean R@1 over the top-100 object-region heads and over the top-10.}
\label{tab:object_region}
\begin{tabular}{l c c c c c c}
\toprule
& \multicolumn{2}{c}{Ranking agreement} & \multicolumn{4}{c}{Same-category retrieval} \\
\cmidrule(lr){2-3} \cmidrule(lr){4-7}
Model & object vs.\ full & control $\rho$ / overlap & $m$ & chance & top-100 R@1 & top-10 R@1 \\
\midrule
InternVL3.5-4B & 0.970 & 0.993 / 93 & 12.1 & 0.091 & 0.466 & 0.592 \\
InternVL3.5-8B & 0.970 & 0.993 / 96 & 14.2 & 0.077 & 0.424 & 0.578 \\
Qwen3-VL-4B & 0.971 & 0.996 / 96 & 16.4 & 0.064 & 0.485 & 0.633 \\
Qwen3-VL-8B & 0.970 & 0.996 / 96 & 16.4 & 0.064 & 0.429 & 0.590 \\
\bottomrule
\end{tabular}
\end{table}

The object-region ranking agrees with the full-image writer ranking at $\rho = 0.970$ to $0.971$, somewhat below the agreement between two full-image trainings ($\rho \ge 0.993$, 93 to 96 shared top-100 heads), so object information changes the ranking by more than training noise but leaves it largely intact.
On the same-category test the top-100 object-region heads retrieve the correct region in 42 to 48\% of queries against a chance of 6 to 9\%, and the top-10 heads in 58 to 63\%.
Writer heads therefore carry information about the queried object itself, not only about which image it came from.

\subsection{In-context candidates instead of a held-out pool}
The writer criterion scores each head against a held-out pool of 1,200 candidate images.
A natural alternative scores it only against the images of the same context, five or six candidates, which is closer to the retrieval the model actually performs.
Table~\ref{tab:incontext} compares this in-context ranking with the held-out writer ranking and with the pointer ranking, using a paired head bootstrap (5,000 resamples) for the difference of the two correlations, and a split-half diagnostic in which the test samples are divided into two halves, heads are ranked on each half, and we count the top-100 heads shared by the two halves.

\begin{table}[h]
\centering
\footnotesize
\caption{In-context writer ranking versus the held-out writer and pointer rankings (Spearman $\rho$ over 1,152 heads; $\Delta = \rho_{\mathrm{writer}} - \rho_{\mathrm{pointer}}$ with its 95\% bootstrap CI), and split-half stability of the two writer scorings: top-100 heads shared between halves and the R@1 gap between the rank-1 and rank-100 head (points).}
\label{tab:incontext}
\setlength{\tabcolsep}{3pt}
\begin{tabular}{l c c c c @{\hspace{0.8em}} c c c c}
\toprule
& \multicolumn{4}{c}{Agreement of the in-context ranking} & \multicolumn{4}{c}{Split-half stability} \\
\cmidrule(lr){2-5} \cmidrule(lr){6-9}
& & & & & \multicolumn{2}{c}{shared top-100} & \multicolumn{2}{c}{rank 1 vs.\ 100 gap} \\
Model & vs.\ writer & vs.\ pointer & $\Delta$ & 95\% CI & in-ctx & held-out & in-ctx & held-out \\
\midrule
InternVL3.5-4B & 0.758 & 0.480 & 0.277 & [0.231, 0.325] &  65 & 85 & 4.3 & 22.0 \\
InternVL3.5-8B & 0.716 & 0.530 & 0.186 & [0.143, 0.231] &  71 & 87 & 4.0 & 24.4 \\
Qwen3-VL-4B & 0.573 & 0.403 & 0.170 & [0.124, 0.217] &  61 & 86 & 1.8 & 19.5 \\
Qwen3-VL-8B & 0.657 & 0.406 & 0.251 & [0.202, 0.302] &  63 & 84 & 2.2 & 28.9 \\
\bottomrule
\end{tabular}
\end{table}

The in-context ranking is closer to the held-out writer ranking than to the pointer ranking in every model, with the difference bounded away from zero.
It is, however, a poorly resolved criterion: with only a handful of candidates most heads score near the ceiling, the rank-1 and rank-100 heads differ by only 1.8 to 4.3 points, and the two halves of the test set agree on just 61 to 71 of their top-100 heads, against 84 to 87 for the held-out scoring whose rank-1 to rank-100 gap is about 20 to 29 points.
The held-out pool is used in the main text for this reason.

\section{Where Writer-Only Heads Get Their Information}
\label{app:source_decomp}

Writer-only heads are a relative partition: heads in the writer top-100 but outside the pointer top-100 (\S\ref{sec:ablation}).
The probe recovers the target image from their outputs although they fall outside the pointer top-100, which raises the question of where that information comes from: from the target-image tokens themselves, or relayed through text positions.
Following Eq.~\ref{eq:head_output}, we decompose each writer-only head's output at the cattok into the contributions of four source groups: target-image tokens, distractor-image tokens, interleaved filler passages, and the remaining question and option tokens.
We then apply the writer probe of \S\ref{sec:writer}, unchanged, to the output reconstructed from one source group alone or with one group removed, and report the mean writer R@1 over the writer-only heads of each model (Table~\ref{tab:source_decomp}).
Because the probe and the cosine ranking are nonlinear, a single source can score slightly above the full output; the rows are sufficiency and removal tests, not an additive decomposition.
The two removal rows carry 95\% confidence intervals from 5,000 paired head-bootstrap resamples within each model's writer-only set; both exclude zero in every model.

\begin{table}[h]
\centering
\footnotesize
\setlength{\tabcolsep}{4pt}
\caption{Writer R@1 of writer-only heads (mean over the set) when the head output is reconstructed from one source group or with one group removed. The two removal rows give the change relative to the full output, with its 95\% bootstrap CI beneath. Text tokens are the interleaved filler passages plus the question and option tokens (including special tokens).}
\label{tab:source_decomp}
\begin{tabular}{l c c c c}
\toprule
Source intervention & InternVL3.5-4B & InternVL3.5-8B & Qwen3-VL-4B & Qwen3-VL-8B \\
\midrule
Full head output & 0.331 & 0.285 & 0.352 & 0.346 \\
Target-image tokens only & 0.338 & 0.341 & 0.322 & 0.326 \\
Remove target-image tokens ($\Delta$) & $-0.175$ & $-0.160$ & $-0.163$ & $-0.167$ \\
\quad 95\% CI & [$-0.198$, $-0.150$] & [$-0.182$, $-0.138$] & [$-0.185$, $-0.140$] & [$-0.189$, $-0.144$] \\
Text tokens only & 0.155 & 0.126 & 0.190 & 0.178 \\
Remove all text tokens ($\Delta$) & $-0.049$ & $-0.025$ & $-0.085$ & $-0.075$ \\
\quad 95\% CI & [$-0.075$, $-0.025$] & [$-0.050$, $-0.001$] & [$-0.112$, $-0.061$] & [$-0.101$, $-0.051$] \\
Interleaved filler only & 0.003 & 0.003 & 0.002 & 0.003 \\
Distractor-image tokens only & 0.004 & 0.003 & 0.002 & 0.002 \\
\bottomrule
\end{tabular}
\end{table}

The target-image tokens alone retain a writer R@1 comparable to the full output, and removing them causes the largest drop in every model.
Text tokens alone stay well below the full output, and removing all text costs much less than removing the target-image tokens; interleaved filler and distractor-image tokens alone carry almost nothing.
Text therefore provides secondary information, but a relay through text positions is not required for the target-image information that the probe recovers.

\paragraph{Attention patterns and pointer ranks.}
The attention patterns point in the same direction.
At the cattok and at positions $+1$ to $+4$, writer-only heads place 2.1 to 4.0 times as much attention mass on the target image as layer-matched random heads, while their attention to the interleaved filler is below that of random heads (0.74 to 0.89 times).
Nor do they become strong pointers at a neighboring position: across $+1$ to $+4$ the median pointer rank of writer-only heads stays above 200 of 1,152, and at none of these positions in any model do more than 12 of them enter the pointer top-100.
Attention magnitude alone also does not determine output content: among heads with similar pointer R@1 (0.30 to 0.40), writer R@1 differs by 4.0 to 4.2 times between the 10th and 90th percentiles, consistent with Eq.~\ref{eq:head_output}, in which attention weights set how sources are mixed while the value pathway sets what the mixture carries~\citep{kobayashi2020attention}.

\paragraph{Scope.}
The decomposition localizes the source positions of the current head's output; it does not say how the target-image-token representations formed in earlier layers.
Writer R@1 measures probe recoverability, not downstream use: writer-only ablations remain near the matched-budget random-head controls (\S\ref{sec:ablation}, Appendix~\ref{app:set_partition}).

\section{Readout Position: Both Signals Are Anchored at the Cattok}
\label{app:position}

The pointer and writer criteria are read at the cattok, the last token of the queried-category mention (\S\ref{sec:cocomcq}).
This appendix checks how much of each signal survives when the readout moves to later positions: the four tokens following the cattok ($+1$ to $+4$), the token midway between the cattok and the end of the prompt (mid), and the last prefill token (last).
The pointer ranking is recomputed at every position from the attention mass of that position over each image span (Eq.~\ref{eq:pointer_score}) on the 6,000 ICIR-MCQ samples.
The writer probes are retrained independently at each position ($+1$, $+2$, $+4$, mid, last) with the configuration of Appendix~\ref{app:learned_retriever}; a probe retrained at the cattok itself reproduces the cattok writer ranking at Spearman $\rho = 0.993$ on every model, which serves as the reference for retraining variability.
Table~\ref{tab:position_control} reports, for each position, the Spearman correlation of the full 1,152-head ranking with the cattok ranking together with the top-100 overlap, and the retained signal of the head sets fixed at the cattok: the mean R@1 of the cattok pointer top-100 (writer top-100) at that position divided by its value at the cattok.

\begin{table}[h]
\centering
\footnotesize
\setlength{\tabcolsep}{3pt}
\caption{Readout-position control. First row per model: Spearman $\rho$ of the pointer ranking at each position against the cattok pointer ranking, and the top-100 overlap. Second row: mean pointer R@1 of the pointer top-100 of this sweep's cattok ranking, at each position divided by its value at the cattok. Third row: the same correlation and overlap for the writer ranking retrained at each position; the cattok entry is the retrained cattok probe against the original cattok ranking. Fourth row: mean writer R@1 of the fixed cattok writer top-100 under the probe retrained at each position, divided by its value under the retrained cattok probe. Writer probes were not retrained at $+3$.}
\label{tab:position_control}
\begin{tabular}{l c c c c c c c}
\toprule
& cattok & $+1$ & $+2$ & $+3$ & $+4$ & mid & last \\
\midrule
InternVL3.5-4B & -- & 0.960 / 79 & 0.946 / 77 & 0.945 / 75 & 0.948 / 76 & 0.957 / 78 & 0.683 / 52 \\
\quad pointer set & 1.00 & 0.86 & 0.81 & 0.80 & 0.86 & 0.82 & 0.34 \\
\quad writer ranking $\rho$ / overlap & 0.993 / -- & 0.850 / 72 & 0.812 / 67 & -- & 0.847 / 70 & 0.810 / 65 & 0.536 / 16 \\
\quad writer set & 1.00 & 0.73 & 0.57 & -- & 0.68 & 0.51 & 0.32 \\
\addlinespace[1pt]
InternVL3.5-8B & -- & 0.957 / 77 & 0.949 / 74 & 0.948 / 73 & 0.943 / 72 & 0.959 / 73 & 0.682 / 50 \\
\quad pointer set & 1.00 & 0.85 & 0.81 & 0.80 & 0.85 & 0.84 & 0.41 \\
\quad writer ranking $\rho$ / overlap & 0.993 / -- & 0.847 / 63 & 0.792 / 62 & -- & 0.823 / 59 & 0.818 / 57 & 0.558 / 17 \\
\quad writer set & 1.00 & 0.69 & 0.54 & -- & 0.62 & 0.51 & 0.46 \\
\addlinespace[1pt]
Qwen3-VL-4B & -- & 0.961 / 87 & 0.949 / 81 & 0.946 / 82 & 0.945 / 85 & 0.959 / 82 & 0.739 / 51 \\
\quad pointer set & 1.00 & 0.84 & 0.77 & 0.78 & 0.85 & 0.79 & 0.60 \\
\quad writer ranking $\rho$ / overlap & 0.993 / -- & 0.878 / 67 & 0.820 / 59 & -- & 0.857 / 60 & 0.803 / 53 & 0.713 / 26 \\
\quad writer set & 1.00 & 0.63 & 0.44 & -- & 0.55 & 0.36 & 0.82 \\
\addlinespace[1pt]
Qwen3-VL-8B & -- & 0.959 / 87 & 0.943 / 85 & 0.942 / 82 & 0.939 / 86 & 0.954 / 80 & 0.727 / 60 \\
\quad pointer set & 1.00 & 0.85 & 0.77 & 0.77 & 0.85 & 0.78 & 0.61 \\
\quad writer ranking $\rho$ / overlap & 0.993 / -- & 0.875 / 67 & 0.829 / 64 & -- & 0.851 / 68 & 0.815 / 57 & 0.730 / 33 \\
\quad writer set & 1.00 & 0.67 & 0.46 & -- & 0.56 & 0.39 & 0.98 \\
\addlinespace[1pt]
\bottomrule
\end{tabular}
\end{table}

Both rankings stay close to their cattok version at the local positions and at mid: pointer $\rho$ 0.939--0.961 with top-100 overlap 72--87, writer $\rho$ 0.79--0.88, and the top-1 pointer head is unchanged at every local position in every model.
The fixed sets keep a substantial fraction of their signal there: at $+1$ to $+4$ the pointer set retains 0.77--0.86 of its cattok R@1 and the writer set 0.44--0.73.
At the last prefill token both rankings move further from the cattok ranking (pointer $\rho$ 0.68--0.74, writer $\rho$ 0.54--0.73), but both signals remain measurable.
The discovered head populations therefore persist beyond the single scoring position, while the cattok remains the position at which both signals are strongest.

\section{Majority-vote retrieval with the writer ranking}
\label{app:retriever_mv}

This appendix asks how the writer ranking transfers to Visual Haystacks~\citep{wu2025visual}, and whether ensembling the top-$K$ writer-ranked heads recovers the gap to the single best pointer head.
We pair each ranking with the scoring rule it was ranked under, so the comparison is between matched ranking--metric pairs rather than between two metrics applied to a fixed head set.

\begin{figure}[!htbp]
\centering
\includegraphics[width=\linewidth]{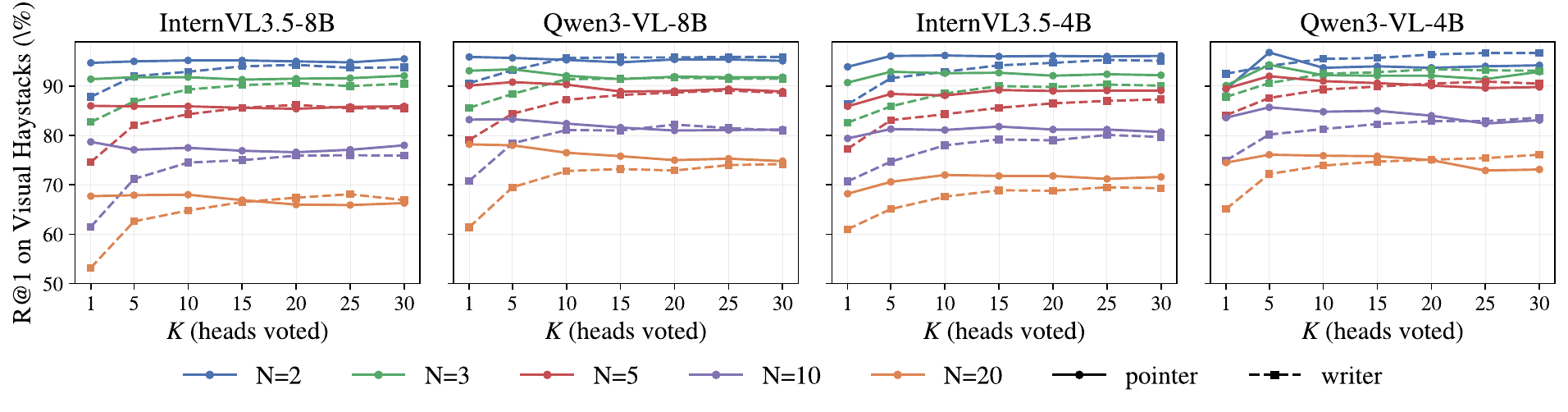}
\caption{Top-$K$ majority-vote R@1 on Visual Haystacks under the two matched ranking--metric pairs, for all four LVLMs ($n{=}1{,}000$ per haystack size). Solid lines: pointer ranking, attention-score voting (A). Dashed lines: writer ranking, trained-probe voting (B). Each color is one haystack size $\in\{2,3,5,10,20\}$.}
\label{fig:majority_vote_kvsweep}
\end{figure}

\paragraph{Head-level vote and majority-vote scoring.}
Let $\mathcal{H}^{\mathrm{ptr}}_K$ and $\mathcal{H}^{\mathrm{wri}}_K$ denote the top-$K$ heads under the pointer and writer rankings.
For a VHs sample with image set $\{e_i\}_{i=1}^{N}$:
\begin{itemize}
    \item Pointer-ranking, attention-score voting (A): each $h \in \mathcal{H}^{\mathrm{ptr}}_K$ casts a vote $v_h^A {=} \arg\max_i \, a_{h,i}$, where $a_{h,i}$ is the cattok attention sum from head $h$ to image $i$'s patches.
\item Writer-ranking, trained-probe voting (B): each 
$h \in \mathcal{H}^{\mathrm{wri}}_K$ casts a vote
\[
v_h^B
=
\arg\max_i 
\left\langle 
\tilde{f}_h(o_h),\,
\tilde{g}_h(e_i)
\right\rangle ,
\]
where $f_h$ and $g_h$ are the trained head-side and image-side projectors of head $h$ (Appendix~\ref{app:learned_retriever}), and $\tilde{\cdot}$ denotes $\ell_2$ normalization.
\end{itemize}
The sample-level prediction is the majority vote over the $K$ votes (with ties broken by smallest image index); $K=1$ recovers the standalone single-head predictor in each case.
We report R@1 against the ground-truth target image position on $1{,}000$ samples per haystack size for all four LVLMs; Figure~\ref{fig:majority_vote_kvsweep} plots the per-LVLM dose-response.

\paragraph{Single head: the pointer ranking is stronger in 19 of 20 cells.}
At $K{=}1$ the pointer top head outperforms the writer top head in 19 of 20 (model, $N$) cells, by $2.3$ to $17.2$\,pp; the exception is Qwen3-VL-4B at $N{=}2$ ($89.4$ vs.\ $92.5$). The gap widens with haystack size, to about $17$\,pp on the 8B models and about $9$\,pp on the 4B models.
The pointer ranking thus identifies the stronger individual head, even though the writer probe was trained for retrieval.

\paragraph{Ensembling closes most of the single-head gap.}
Majority vote over the top-$K$ writer-probe heads improves with $K$ and saturates by $K \approx 20$--$30$ on all four models; at saturation it is within $4.0$\,pp of the best pointer configuration in every (model, $N$) cell, within $1.2$\,pp at $N{=}2$, and matches or marginally exceeds it in four cells.
The pointer ranking behaves differently, and the difference depends on scale: on the 8B models a single head is already at or near the best (adding heads gains at most $0.8$\,pp), while on the 4B models a small ensemble helps (up to $+7.4$\,pp at $K{=}5$ on Qwen3-VL-4B).
The pointer ranking thus identifies a few heads whose raw attention pattern is by itself an accurate retriever, whereas the writer ranking identifies heads whose outputs carry retrieval information that is realized in aggregate rather than head by head.

\section{Set-Partition Ablation: overlap vs.\ pointer-only vs.\ writer-only}
\label{app:set_partition}

\begin{figure}[h]
\centering
\includegraphics[width=\linewidth]{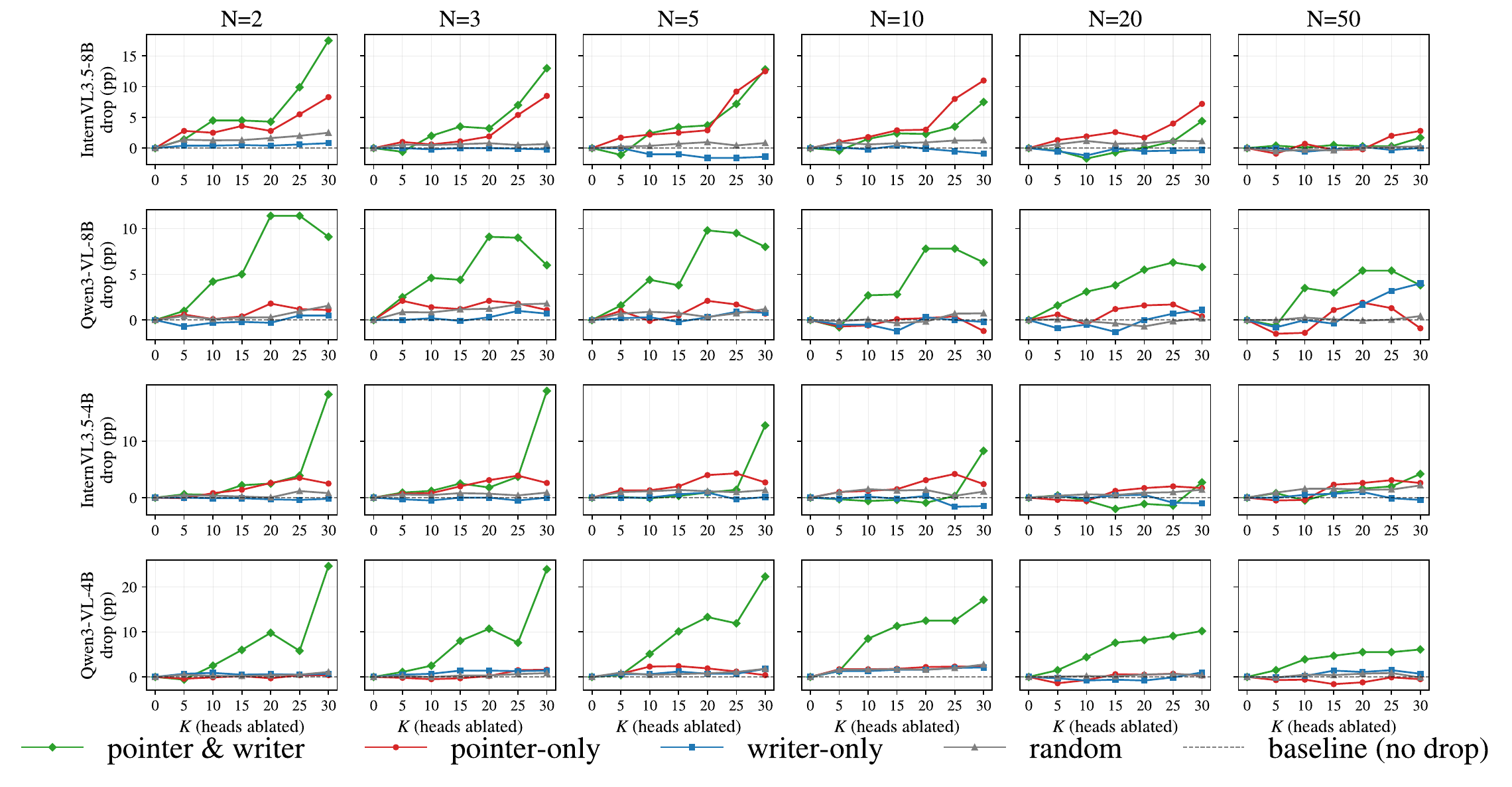}
\caption{Set-partition dose-response per LVLM (top-100 partition, output-zero). Rows: LVLMs (top to bottom: InternVL3.5-8B, Qwen3-VL-8B, InternVL3.5-4B, Qwen3-VL-4B). Columns: haystack size $\in\{2,3,5,10,20,50\}$. Random is the mean over 3 seeds.}
\label{fig:setpart_per_model_grid}
\end{figure}

To isolate which part of the pointer top-100 and writer top-100 drives the causal effect, we partition their union into three disjoint subsets and run the targeted output-zero ablation of \S\ref{sec:validation} (Appendix~\ref{app:impl}) on each at $K \in \{5, 10, 15, 20, 25, 30\}$ and $N \in \{2,3,5,10,20,50\}$.
Within each subset heads are ordered by pointer rank (overlap, pointer-only) or writer rank (writer-only), and the first $K$ heads of that order are ablated; the overlap subset has 35--47 heads and pointer-only and writer-only 53--65 each, depending on the model.
A random control of the same $K$ uniformly sampled heads serves as the reference.
Figure~\ref{fig:setpart_per_model_grid} plots the per-model dose-response curves.

Table~\ref{tab:overlap_heads} lists the overlap heads of each model in ablation order. The pointer top-1 head \texttt{L21H10} is in the overlap of every model (writer rank 35/33/55/50 on InternVL3.5-8B / InternVL3.5-4B / Qwen3-VL-8B / Qwen3-VL-4B), and 23 heads are in the overlap of all four models.

\begin{table}[h]
\centering
\footnotesize
\caption{Overlap heads (in both pointer top-100 and writer top-100, cattok query) per LVLM, listed by pointer rank, the order in which the overlap subset is ablated (the first $K$ heads form the budget-$K$ overlap arm). Format \texttt{LxxHyy}(a$x$/r$y$) gives the head's 1-indexed pointer rank $x$ and writer rank $y$.}
\label{tab:overlap_heads}
\begin{tabularx}{\linewidth}{@{}l X@{}}
\toprule
Model & Overlap heads (pointer-rank order) \\
\midrule
InternVL3.5-8B (47 heads) & \texttt{L21H10}(a1/r35), \texttt{L21H11}(a2/r46), \texttt{L21H08}(a3/r20), \texttt{L20H17}(a4/r83), \texttt{L21H18}(a7/r10), \texttt{L20H16}(a8/r81), \texttt{L20H31}(a9/r16), \texttt{L20H21}(a10/r68), \texttt{L21H17}(a11/r6), \texttt{L20H15}(a14/r3), \texttt{L19H24}(a17/r97), \texttt{L20H30}(a18/r8), \texttt{L22H04}(a20/r34), \texttt{L23H04}(a21/r82), \texttt{L20H28}(a23/r2), \texttt{L21H31}(a24/r4), \texttt{L20H13}(a25/r63), \texttt{L24H29}(a27/r14), \texttt{L24H05}(a28/r7), \texttt{L18H19}(a29/r5), \texttt{L20H14}(a32/r45), \texttt{L19H27}(a34/r42), \texttt{L24H23}(a36/r1), \texttt{L19H13}(a38/r25), \texttt{L28H23}(a41/r91), \texttt{L19H03}(a42/r94), \texttt{L17H04}(a46/r51), \texttt{L22H06}(a49/r22), \texttt{L23H31}(a50/r90), \texttt{L28H02}(a51/r69), \texttt{L13H11}(a52/r38), \texttt{L19H28}(a54/r49), \texttt{L19H14}(a58/r74), \texttt{L18H18}(a59/r9), \texttt{L24H13}(a64/r33), \texttt{L19H25}(a69/r75), \texttt{L31H30}(a73/r96), \texttt{L23H09}(a78/r27), \texttt{L29H15}(a79/r64), \texttt{L19H12}(a82/r44), \texttt{L29H11}(a85/r77), \texttt{L15H11}(a87/r24), \texttt{L23H11}(a89/r17), \texttt{L23H28}(a91/r18), \texttt{L20H29}(a94/r99), \texttt{L18H05}(a99/r47), \texttt{L24H14}(a100/r19) \\
\addlinespace[2pt]
Qwen3-VL-8B (35 heads) & \texttt{L21H10}(a1/r55), \texttt{L21H08}(a2/r21), \texttt{L21H11}(a3/r60), \texttt{L21H18}(a6/r18), \texttt{L20H15}(a9/r24), \texttt{L19H27}(a10/r16), \texttt{L21H17}(a11/r2), \texttt{L20H31}(a13/r28), \texttt{L22H04}(a14/r86), \texttt{L20H30}(a17/r6), \texttt{L18H19}(a23/r7), \texttt{L21H31}(a24/r4), \texttt{L24H29}(a25/r10), \texttt{L19H13}(a28/r96), \texttt{L20H28}(a30/r5), \texttt{L24H05}(a32/r11), \texttt{L17H04}(a36/r45), \texttt{L24H23}(a37/r1), \texttt{L18H18}(a40/r17), \texttt{L22H06}(a45/r80), \texttt{L28H02}(a46/r26), \texttt{L24H30}(a49/r58), \texttt{L23H30}(a54/r68), \texttt{L25H09}(a62/r81), \texttt{L24H13}(a67/r15), \texttt{L19H25}(a68/r91), \texttt{L18H05}(a70/r49), \texttt{L30H06}(a71/r25), \texttt{L26H31}(a75/r29), \texttt{L25H11}(a84/r76), \texttt{L23H28}(a90/r62), \texttt{L17H27}(a91/r79), \texttt{L23H12}(a93/r97), \texttt{L28H03}(a97/r43), \texttt{L24H14}(a98/r8) \\
\addlinespace[2pt]
InternVL3.5-4B (42 heads) & \texttt{L21H10}(a1/r33), \texttt{L21H11}(a2/r23), \texttt{L21H08}(a4/r29), \texttt{L21H18}(a6/r9), \texttt{L23H04}(a10/r92), \texttt{L24H29}(a11/r8), \texttt{L21H17}(a12/r5), \texttt{L20H31}(a13/r27), \texttt{L21H16}(a14/r89), \texttt{L20H16}(a16/r91), \texttt{L20H15}(a17/r4), \texttt{L21H23}(a21/r59), \texttt{L21H31}(a22/r1), \texttt{L28H23}(a23/r63), \texttt{L22H04}(a25/r24), \texttt{L20H30}(a26/r6), \texttt{L24H23}(a29/r3), \texttt{L13H11}(a30/r100), \texttt{L24H05}(a32/r15), \texttt{L20H18}(a33/r74), \texttt{L20H14}(a36/r73), \texttt{L20H28}(a38/r2), \texttt{L28H03}(a39/r97), \texttt{L31H30}(a40/r85), \texttt{L28H02}(a41/r21), \texttt{L19H27}(a45/r50), \texttt{L24H13}(a46/r53), \texttt{L18H19}(a47/r11), \texttt{L24H14}(a49/r22), \texttt{L17H04}(a51/r66), \texttt{L23H28}(a52/r13), \texttt{L19H13}(a53/r36), \texttt{L29H13}(a55/r99), \texttt{L22H06}(a61/r38), \texttt{L21H30}(a69/r84), \texttt{L19H14}(a79/r88), \texttt{L15H11}(a84/r96), \texttt{L19H28}(a86/r60), \texttt{L18H18}(a87/r18), \texttt{L19H12}(a88/r82), \texttt{L28H16}(a99/r98), \texttt{L26H18}(a100/r20) \\
\addlinespace[2pt]
Qwen3-VL-4B (43 heads) & \texttt{L21H10}(a1/r50), \texttt{L21H11}(a3/r85), \texttt{L21H08}(a4/r28), \texttt{L21H18}(a6/r15), \texttt{L22H04}(a8/r55), \texttt{L21H17}(a9/r4), \texttt{L20H16}(a10/r62), \texttt{L20H31}(a11/r8), \texttt{L20H15}(a12/r12), \texttt{L20H17}(a13/r77), \texttt{L19H24}(a14/r69), \texttt{L19H27}(a17/r19), \texttt{L20H18}(a18/r58), \texttt{L21H31}(a19/r3), \texttt{L21H23}(a22/r86), \texttt{L18H19}(a23/r5), \texttt{L19H13}(a24/r48), \texttt{L20H30}(a25/r7), \texttt{L24H29}(a26/r26), \texttt{L28H23}(a29/r87), \texttt{L19H14}(a30/r36), \texttt{L19H12}(a35/r53), \texttt{L21H30}(a38/r82), \texttt{L20H28}(a40/r2), \texttt{L30H06}(a42/r40), \texttt{L22H06}(a45/r32), \texttt{L18H18}(a46/r6), \texttt{L24H23}(a51/r1), \texttt{L18H05}(a56/r33), \texttt{L17H04}(a58/r31), \texttt{L24H13}(a61/r59), \texttt{L24H05}(a62/r43), \texttt{L19H28}(a67/r63), \texttt{L19H25}(a72/r49), \texttt{L28H02}(a76/r72), \texttt{L24H30}(a77/r92), \texttt{L13H11}(a83/r25), \texttt{L17H05}(a84/r95), \texttt{L24H14}(a87/r16), \texttt{L17H27}(a91/r57), \texttt{L23H12}(a92/r64), \texttt{L23H14}(a98/r80), \texttt{L26H31}(a100/r71) \\
\addlinespace[2pt]
\bottomrule
\end{tabularx}
\end{table}

\subsection{Prefill-only versus full-generation intervention}
\label{app:gen_window}
The ablations above stop at the last prefill position, so the first Yes/No logit already reflects the intervention while the generated tokens are untouched (Appendix~\ref{app:impl}).
To check that stopping at the prefill does not change the conclusions, we re-ran the three arms and the three random controls at $N = 10$, $K = 30$ with the zeroing continued through every generated token, and compared the two runs sample by sample (Table~\ref{tab:gen_window}).
For both InternVL models, which answer with a single word, no prediction changes.
For the Qwen3-VL models, whose answers can run to a full sentence (mostly on the 8B model), continuing the intervention touches tokens after the decision and changes at most 5.1\% of parsed answers, moving accuracy by at most 0.9 points without changing the ordering of the arms.

\begin{table}[h]
\centering
\small
\caption{Full-generation minus prefill-only intervention at $N = 10$, $K = 30$ (VHs, $n = 1{,}000$): change in accuracy (points) and fraction of samples whose parsed answer changed.}
\label{tab:gen_window}
\begin{tabular}{l c c c c}
\toprule
& overlap & pointer-only & writer-only & random (3 seeds) \\
\midrule
InternVL3.5-4B & 0.0 (0\%) & 0.0 (0\%) & 0.0 (0\%) & 0.0 (0\%) \\
InternVL3.5-8B & 0.0 (0\%) & 0.0 (0\%) & 0.0 (0\%) & 0.0 (0\%) \\
Qwen3-VL-4B & 0.0 (0\%) & 0.0 (0\%) & 0.0 (0\%) & $+0.1$ / $+0.1$ / $-0.1$ (0.1--0.3\%) \\
Qwen3-VL-8B & $-0.4$ (4.1\%) & $-0.8$ (5.1\%) & $-0.9$ (2.2\%) & $-0.2$ / $+0.8$ / $0.0$ (0--1.7\%) \\
\bottomrule
\end{tabular}
\end{table}

\section{Replication on Additional LVLMs}
\label{app:replication}

\subsection{Discovery on three further models}
We applied the pointer and writer criteria, unchanged, to three additional LVLMs: LLaVA-OneVision-7B~\citep{li2024llava} (Qwen2 backbone, SigLIP vision tower), Pixtral-12B~\citep{agrawal2024pixtral} (Mistral-Nemo backbone, Pixtral vision encoder), and InternVL3.5-14B~\citep{wang2025internvl3} (a larger member of the InternVL family).
Probes use the configuration of Appendix~\ref{app:learned_retriever}.
Images are processed at a fixed size per model so that candidate images contribute equal numbers of visual tokens: 384-pixel single-tile inputs for LLaVA-OneVision, whose any-resolution default would otherwise give images different token counts, and 448 pixels for Pixtral.
Table~\ref{tab:replication_discovery} reports the pointer top-1 head R@1 on ICIR-MCQ and the pointer-writer overlap among the top-100 heads, together with the overlap expected if the two rankings were unrelated ($100^2/H$ for $H$ heads).

\begin{table}[h]
\centering
\small
\caption{Pointer-writer discovery on additional LVLMs (ICIR-MCQ, 6,000 samples). Overlap counts heads in both top-100 lists; chance is $100^2/H$.}
\label{tab:replication_discovery}
\setlength{\tabcolsep}{3pt}
\begin{tabular}{l l r c c c c}
\toprule
Model & LLM / vision & $H$ & Top-1 R@1 & Overlap & Chance & Ratio \\
\midrule
Main-text models (4) & InternVL3.5, Qwen3-VL & 1,152 & 0.866--0.922 & 35--47 & 8.7 & 4.0--5.4$\times$ \\
LLaVA-OneVision-7B & Qwen2 / SigLIP & 784 & 0.962 & 46 & 12.8 & 3.6$\times$ \\
Pixtral-12B & Mistral-Nemo / Pixtral & 1,280 & 0.874 & 40 & 7.8 & 5.1$\times$ \\
InternVL3.5-14B & Qwen3-14B / InternViT & 1,600 & 0.844 & 31 & 6.2 & 5.0$\times$ \\
\bottomrule
\end{tabular}
\end{table}

All three models reproduce the discovery-side pattern: a single pointer head retrieves the target image with high R@1, and the pointer and writer top-100 lists overlap several times more than chance while remaining far from identical.

\subsection{Set-partition ablation on LLaVA-OneVision-7B}
We ran the complete set-partition protocol of \S\ref{sec:validation} on LLaVA-OneVision-7B under the same conditions as the four main-text models: Visual Haystacks, $n = 1{,}000$ per haystack size, the output-zero intervention of \S\ref{sec:validation}, top-100 partition, $K \in \{5, 10, 15, 20, 25, 30\}$, and matched-budget random controls over three seeds.
Table~\ref{tab:llava_causal} reports accuracy relative to the random control at each $(N, K)$.
Haystack size 50 is omitted for this model: with about 1,485 visual tokens per image the $N{=}50$ prompt runs to about 74k tokens, more than twice the model's 32,768-token training context, and the unablated baseline itself falls below chance, so no intervention effect can be measured there. The four main-text models use at most 256 tokens per image (Appendix~\ref{app:impl}) and stay within context.

\begin{table}[h]
\centering
\small
\setlength{\tabcolsep}{4pt}
\caption{LLaVA-OneVision-7B set-partition ablation: accuracy minus the matched random control (mean of three seeds) for the pointer-and-writer overlap (first row per $N$), pointer-only, and writer-only sets. Baseline is the unablated accuracy at each haystack size.}
\label{tab:llava_causal}
\begin{tabular}{l c c c c c c c}
\toprule
$N$ & Baseline & $K{=}5$ & $K{=}10$ & $K{=}15$ & $K{=}20$ & $K{=}25$ & $K{=}30$ \\
\midrule
2 & 0.761 & $-0.123$ & $-0.180$ & $-0.198$ & $-0.209$ & $-0.211$ & $-0.213$ \\
\quad pointer-only & & $-0.011$ & $-0.094$ & $-0.098$ & $-0.100$ & $-0.106$ & $-0.132$ \\
\quad writer-only & & $-0.003$ & $-0.005$ & $-0.002$ & $-0.007$ & $-0.009$ & $-0.017$ \\
\addlinespace[1pt]
3 & 0.741 & $-0.144$ & $-0.193$ & $-0.207$ & $-0.213$ & $-0.214$ & $-0.210$ \\
\quad pointer-only & & $-0.007$ & $-0.083$ & $-0.114$ & $-0.113$ & $-0.113$ & $-0.136$ \\
\quad writer-only & & $-0.004$ & $-0.003$ & $-0.007$ & $-0.007$ & $-0.015$ & $-0.017$ \\
\addlinespace[1pt]
5 & 0.685 & $-0.100$ & $-0.165$ & $-0.167$ & $-0.160$ & $-0.149$ & $-0.156$ \\
\quad pointer-only & & $-0.005$ & $-0.055$ & $-0.080$ & $-0.063$ & $-0.054$ & $-0.075$ \\
\quad writer-only & & $+0.003$ & $-0.002$ & $-0.007$ & $-0.005$ & $+0.005$ & $-0.005$ \\
\addlinespace[1pt]
10 & 0.637 & $-0.069$ & $-0.105$ & $-0.112$ & $-0.094$ & $-0.083$ & $-0.088$ \\
\quad pointer-only & & $-0.013$ & $-0.028$ & $-0.041$ & $-0.017$ & $-0.010$ & $-0.017$ \\
\quad writer-only & & $+0.013$ & $+0.014$ & $+0.006$ & $+0.013$ & $+0.015$ & $+0.010$ \\
\addlinespace[1pt]
20 & 0.577 & $-0.041$ & $-0.067$ & $-0.066$ & $-0.056$ & $-0.047$ & $-0.047$ \\
\quad pointer-only & & $-0.016$ & $-0.015$ & $-0.003$ & $+0.011$ & $+0.008$ & $-0.014$ \\
\quad writer-only & & $+0.006$ & $+0.008$ & $+0.003$ & $+0.003$ & $+0.020$ & $+0.018$ \\
\addlinespace[1pt]
\bottomrule
\end{tabular}
\end{table}

The overlap set is the most damaging arm in all 30 cells, and its drop is significant (McNemar, $p<0.05$) against all three random seeds in 24 of the 30 cells and against at least two seeds in all 30.
The sparsity of the effect is visible at $K=5$: removing five overlap heads already costs 0.07 to 0.14 accuracy at $N \le 10$, while five pointer-only or writer-only heads stay within 0.02 of the random control.
Writer-only ablations stay within $\pm 0.02$ of random throughout.

\section{Visual vs.\ Text Retrieval Heads}
\label{app:modality}

We examined whether the image retrieval heads of Qwen3-VL-8B also serve text retrieval, using a text needle-in-a-haystack task: a numeric needle is embedded in a long interleaved image-text context (the images are filler and none is a retrieval target), and the model is asked to return the number.
Baseline accuracy is $100\%$ on the 291 samples, so head-level differences reflect attention behavior rather than task difficulty; we rank all 1{,}152 heads by their mean attention mass on the needle tokens at the position where the answer is generated.
The two probes are not matched (see below), so we read the result as suggestive only.
Table~\ref{tab:modality_app} shows selected heads and their ranks in image versus text retrieval.

\begin{table}[h]
\centering
\caption{Image vs.\ text retrieval rankings (Qwen3-VL-8B).}
\label{tab:modality_app}
\begin{tabular}{lccc}
\toprule
Head & Image rank & Text rank & Character \\
\midrule
\rowcolor{lightblue}
L21H10 & 1 & 118 & Image-specific \\
\rowcolor{lightblue}
L21H08 & 2 & 67 & Image-specific \\
\rowcolor{lightblue}
L21H11 & 3 & 66 & Image-specific \\
L23H04 & 20 & 10 & Mixed \\
\rowcolor{lightred}
L29H09 & 137 & 1 & Text-specific \\
\rowcolor{lightred}
L29H11 & 221 & 2 & Text-specific \\
\bottomrule
\end{tabular}
\end{table}

The top image retrieval head ranks 118th for text retrieval and the top text retrieval head ranks 137th for image retrieval; no head is in the top~10 for both.
The pointer cluster sits in layers 19--23, whereas the two strongest text retrieval heads sit at layer 29 and the text top-10 spans layers 23--32.

\paragraph{Why the comparison is not precise.}
The two probes differ in candidate granularity (image spans of hundreds of patch tokens versus a few-token needle), in the scoring rule (argmax of attention mass over image spans versus attention to the needle position), in task format, and in query position (the cattok versus the answer position), any of which can move a head's rank for reasons unrelated to modality.
We therefore take the result as evidence that the two retrieval functions are not served by the same top-ranked heads in this network, and leave a matched probe to future work.

\section{Extended Discussion}
\label{app:extended}

\paragraph{Cross-model convergence.}
Head-level roles that recur across independently trained models are more plausibly a property of the architecture and of multi-image training than a per-model artifact: InternVL3.5 and Qwen3-VL are trained by different organizations on different data mixtures, yet the pointer cluster occupies the same mid-block layers of the same 36-block shape on all four models (Appendix~\ref{app:heads}).
This mid-block position matches where language-only transformers implement retrieval and factual lookup~\citep{wu2025retrieval, geva2023dissecting}, which suggests that the pointer-writer cluster is the multimodal analogue at the same structural position; we do not distinguish between these explanations.

\paragraph{Relation to text retrieval heads.}
\citet{wu2025retrieval} identified retrieval heads in text-only LLMs by an attention-pattern criterion.

We expect input-side selection vs.\ output-side encoding to be a useful axis for revisiting text retrieval heads in long-context QA, but leave that to future work.

\clearpage

\end{document}